\documentclass[a4paper,fleqn]{cas-sc}

\usepackage[numbers,sort&compress]{natbib}

\usepackage{amsmath}
\usepackage{amssymb}
\usepackage{mathtools}
\usepackage{bm}
\usepackage{algorithm}
\usepackage{algpseudocode}

\usepackage{graphicx}
\graphicspath{{./figures/}}
\usepackage{booktabs}
\usepackage{siunitx}
\usepackage{placeins}
\usepackage{multirow}
\usepackage{subcaption}

\usepackage{comment}

\begin{document}

\let\WriteBookmarks\relax
\def\floatpagepagefraction{1}
\def\textpagefraction{.001}

\setcounter{topnumber}{3}
\setcounter{bottomnumber}{2}
\setcounter{totalnumber}{5}

\renewcommand{\topfraction}{0.90}
\renewcommand{\bottomfraction}{0.80}
\renewcommand{\textfraction}{0.10}
\renewcommand{\floatpagefraction}{0.75}


\shorttitle{}

\shortauthors{Mohammad Dastranj and Jouni Mattila}

\title[mode=title]{Modular Kinematic Reduction of Closed-Chain Mechanisms Using Path Assembly and Defect Homotopy}


\author[1]{Mohammad Dastranj}[orcid=0009-0005-7116-4496]
\cormark[1]
\ead{mohammad.dastranj@tuni.fi}

\credit{
  Conceptualization,
  Methodology,
  Software,
  Formal analysis,
  Investigation,
  Validation,
  Visualization,
  Writing -- original draft
}

\author[1]{Jouni Mattila}[orcid=0000-0003-1799-4323]
\ead{jouni.mattila@tuni.fi}

\credit{
  Supervision,
  Validation,
  Writing -- review and editing
}


\affiliation[1]{
  organization={Unit of Automation Technology and Mechanical Engineering, Faculty of Engineering and Natural Sciences, Tampere University},
  postcode={33720 Tampere},
  country={Finland}
}

\cortext[1]{Corresponding author}


\begin{abstract}
Closed kinematic chains complicate modular modeling by coupling active and passive coordinates through nonlinear closure constraints. This paper presents a Path-Assembled Closure Differential Mapping (PACDM) framework for modular closure resolution and kinematic reduction. Each closure element compares two ordered transformation paths with common endpoints, with their mismatch expressed through the logarithm on SE(3) and the corresponding Jacobian assembled from local transformation derivatives. Multi-path modules are constructed from a minimal set of pairwise closure elements, while rank-revealing analysis selects locally independent scalar constraints. A defect homotopy recovers closure-consistent passive coordinates from approximate estimates along a feasible and regular continuation path. At regular configurations, implicit differentiation yields the local active-to-passive differential mapping, which is subsequently used in a predictor–corrector continuation procedure for prescribed motion. The framework is evaluated on a seven-degree-of-freedom heavy-duty manipulator containing two-path and three-path closed-chain modules. Comparison with Simscape Multibody yields trajectory root-mean-square errors below \(8.5\times10^{-10}\) rad, while predictor–corrector continuation is approximately 45.8 times faster than applying defect homotopy at every trajectory sample.
\end{abstract}


\begin{NoHyper}
\maketitle
\end{NoHyper}

\section{Introduction}
\label{sec:introduction}
Closed kinematic chains are widely observed in robotics~\cite{boukheddimi2023}, either in parallel~\cite{lilge2024} or hybrid series-parallel robots~\cite{zhang2026,guo2022}. The benefits that the robots with closed kinematic chains offer include the need for less number of actuators~\cite{li2026}, enhanced precision and load-carrying capacity~\cite{tian2025}, greater agility~\cite{shafei2022}, and improved structural stability~\cite{zhang2026}. In legged robots, for example, adopting closed-chain mechanisms in designing the limbs is gaining more academic~\cite{li2026} and industrial attention~\cite{roig2022} as they provide substantially important benefits for this class of robots, such as making the lower limbs lighter and improving their impact-absorption characteristics~\cite{matteis2025}. Another field that demonstrates the high load-carrying capacity characteristics of closed chains is heavy-duty applications and in systems such as excavators~\cite{zhang2020} and hydraulic manipulators~\cite{cheng2023,shen2023}. 

Closed chains, however, come with a price for their benefits, and it is the complexity of their motion equations, specially at higher degrees of freedom~\cite{shafei2022}, and the difficulty in controlling them that follows~\cite{boukheddimi2023,matteis2025}. One approach to resolve complexity is to adopt modularity~\cite{dastranj2025}. Common existing modular approaches span both multibody dynamics and control. Recursive dynamics formulations such as the Articulated Body Algorithm (ABA)~\cite{featherstone2008} and the Decoupled Natural Orthogonal Complement (DeNOC) method~\cite{shah2013}, as well as control frameworks such as Virtual Decomposition Control (VDC)~\cite{zhu2010}, were originally developed for serial or tree-like multibody systems~\cite{kumar2022,chignoli2025,zhang2025,shah2013}. Consequently, their application to systems containing closed kinematic chains generally requires additional treatment that is often tailored to the specific topology of the robot or mechanism under consideration. In particular, within the VDC framework, existing solutions have therefore focused primarily on particular classes of mechanisms. For example, VDC-based formulations have been presented for multibody systems containing triangular closed kinematic chains composed of several passive revolute joints and a single hydraulically or electrically actuated prismatic joint~\cite{petrovic2022,zhang2025,bahari2025}, a configuration commonly encountered in heavy-duty robotic manipulators.

The need to accommodate closed chains is also evident at the level of reusable robot representations and modeling architectures. The Universal Robot Description Format (URDF), despite its widespread use for tree-structured robotic systems, does not natively represent kinematic loops. URDF+ addresses this limitation while preserving the familiar tree-based elements of URDF, adding elements for loop joints and coupling constraints together with an automated parser for closed-chain models~\cite{chignoli2024}. A related need is evident in parallel-robot modeling. Cable-driven parallel robots encompass a range of configurations for which kinematic and dynamic analysis, workspace characterization, trajectory planning, and control require treatment tailored to their parallel structure and cable characteristics~\cite{zarebidoki2022}. Together, these developments illustrate the broader effort to extend reusable modeling descriptions and analysis tools beyond purely tree-structured robotic systems.

More systematic treatments of closed kinematic chains are provided by constraint-embedding formulations, in which local loop constraints are resolved so that a constrained subsystem can be represented using a reduced set of coordinates. Müller developed a systematic modular formulation for parallel mechanisms with hybrid complex limbs formed by serial arrangements of closed loops~\cite{mueller2022}. Within each limb, the constraints associated with the corresponding fundamental cycles are formulated and resolved locally by partitioning the joint variables into independent and dependent coordinates. When an explicit solution of the geometric constraints is unavailable, the resulting differential constraint relations can also be used to recover the required configuration numerically. Chignoli et al. more recently revisited constraint embedding for rigid-body systems with local kinematic loops, introducing generalized joint models and motion and force subspaces for groups of bodies subject to loop constraints while retaining recursive forward- and inverse-dynamics computations~\cite{chignoli2025}.

A closely related modular treatment is provided by the Hybrid Robot Dynamics (HyRoDyn) framework for series-parallel hybrid robots. In its original analytical formulation, the robot is represented as a serial composition of serial and parallel submechanism modules, with each parallel submechanism described by a loop-closure function that maps its independent coordinates to the corresponding spanning-tree coordinates~\cite{kumar2020}. The associated differential relations are then composed modularly and used in the kinematic and dynamic calculations. This formulation allows analytical loop-closure solutions and their derivatives to be stored and reused for recurring submechanism types. A subsequent numerical--analytical extension retains these analytical descriptions for known submechanisms while resolving the loop constraints of submechanisms without available analytical solutions numerically, thereby allowing both forms of closure resolution to coexist within the same modular model~\cite{kumar2022}. The more recent HyRoDyn formulation integrates explicit loop-closure relations, resolved either analytically or numerically, and their derivatives into minimal-coordinate kinematic and dynamic algorithms~\cite{kumar2024}.

Modular formulations have also been developed for parallel mechanisms with actuation redundancy~\cite{rinaldi2025} and for whole-body control of series-parallel hybrid robots using HyRoDyn~\cite{mronga2022}. These applications require the closed-chain kinematics to be resolved before the reduced quantities can be used in modeling or control. Similarly, minimal-coordinate dynamic modeling requires a mapping from augmented or auxiliary coordinates to minimal coordinates when constraint forces are not retained explicitly~\cite{dastranj2026}.

Accordingly, the problem addressed in this work is narrower than the general treatment of multibody systems with closed kinematic chains. Existing approaches already provide modular constraint embedding, analytical and numerical resolution of loop constraints, and reduced-coordinate kinematic and dynamic formulations~\cite{kumar2022,chignoli2025,mueller2022,kumar2020,kumar2024}. Homotopy continuation has likewise been applied to nonlinear kinematic equations of parallel mechanisms~\cite{gallardoalvarado2019}. The objective here is instead to construct, directly from the ordered transformation paths of a closed-chain module, a systematic closure representation and its local coordinate mapping without requiring a mechanism-specific derivation of a minimal set of scalar closure equations. The proposed framework connects this path-assembly construction with rank-revealing constraint selection and defect-homotopy branch acquisition to provide the kinematic quantities required by a downstream modular model.

The first contribution is a path-assembled formulation of the module closure relations and their differential representation. Each elementary closure is constructed from two ordered transformation paths with common initial and terminal frames. Their transformation mismatch is represented through the logarithm on $\mathrm{SE}(3)$, and differentiation of the resulting residual gives the Path-Assembly Closure Differential Mapping (PACDM) from the local transformation derivatives. For a module containing multiple paths, a graph-theoretically minimal set of two-path closure elements is assembled in a common module-coordinate space. Since the resulting raw residual coordinates need not all be independent, rank-revealing analysis of the assembled differential identifies an independent set of scalar closure relations. The construction therefore avoids requiring a separately derived minimal scalar closure model for each mechanism while retaining the independent constraint representation needed for subsequent coordinate reduction.

The second contribution is the integration of this closure construction with a defect-homotopy procedure for acquiring a closure-consistent passive-coordinate configuration. For prescribed active coordinates and an approximate passive-coordinate estimate, the initial path mismatch is incorporated as a defect so that the approximate configuration becomes an exact solution of an artificial closure problem. The defect is then continuously removed, provided that the corresponding continuation path remains feasible and regular. Independent continuation equations are selected from the defected passive-coordinate Jacobian at the artificial starting configuration and retained during the continuation. After the defect has been removed, the complete raw physical closure is verified and the physical closure Jacobian and its ranks are reevaluated before the independent physical relations used for kinematic reduction are selected. In this way, numerical branch acquisition is kept distinct from the subsequent reduction of the physical closure equations.

Finally, at a regular recovered configuration, the selected physical closure Jacobian is partitioned with respect to the active and passive module coordinates. The nonsingular passive block yields, through implicit differentiation, the local mapping from active-coordinate variations to the corresponding passive-coordinate variations and the associated intermediate Jacobian for expressing the complete module kinematics in terms of its active coordinates. These calculations remain local to each closed-chain module, so independent modules can be processed without forming a robot-wide closure problem. During continuous prescribed motion, the same differential mapping is further used as a first-order predictor for the passive coordinates, followed by direct correction of the physical closure equations; defect homotopy is retained as a recovery procedure when direct correction is unsuccessful. The resulting framework therefore provides a common module-level route from path geometry to closure construction, independent-constraint selection, physical-branch acquisition, and the differential quantities required for subsequent modular modeling and control.

The remainder of this paper is structured as follows. Section~\ref{sec:path-assembly} presents the path-assembly formulation of the closed-chain kinematic constraints, including the two-path closure element, the PACDM, and the construction and rank-revealing reduction of multi-path modules. Section~\ref{sec:defect-homotopy} introduces the defect-homotopy procedure used to acquire a closure-consistent passive-coordinate configuration from an approximate initial estimate. Section~\ref{sec:active-to-passive} derives the subsequent kinematic reduction and active-to-passive coordinate mapping and describes its use for continuous trajectory continuation. Section~\ref{sec:simulation} presents the numerical studies and compares the calculated passive coordinates with the corresponding Simscape Multibody results. Finally, Section~\ref{sec:conclusion} concludes the paper and discusses the resulting scope of the proposed framework.

\section{Path-Assembly Formulation of Closed-Chain Kinematic Constraints}
\label{sec:path-assembly}
Consider a multibody system described by a generalized coordinate vector consisting of active coordinates, whose values are independently prescribed, and passive coordinates, whose values are determined by kinematic closure constraints. Let $\boldsymbol{q}$ denote the global generalized coordinate vector, and let the subscripts $a$ and $p$ indicate its active and passive subvectors, respectively. Consequently,
\begin{equation}
    \boldsymbol{q}=\mathrm{col}\left(\boldsymbol{q}_a , \boldsymbol{q}_p \right),
\end{equation}
where $\mathrm{col}(\cdot)$ denotes vertical concatenation of its vector arguments.

To preserve the modular structure of the system, the coordinates are additionally partitioned into module-level subsets associated with individual open-chain or closed-chain components. Each closed-chain module is processed independently. If a closure relation directly couples the passive coordinates of multiple nominal modules, those nominal modules are treated as a single module. For module $m$, a selection matrix $\boldsymbol{P}_m$ extracts the module-level coordinates from the global coordinate set such that
\begin{equation}
    \widehat{\boldsymbol{q}}_m=\boldsymbol{P}_m \boldsymbol{q}=\mathrm{col}\left(\widehat{\boldsymbol{q}}_{m,a} , \widehat{\boldsymbol{q}}_{m,p}\right).
\end{equation}

We assume that the module coordinates and ordered kinematic paths have been identified from the mechanism graph. Each module is represented by a family of paths with common initial and terminal frames, and equality of these path transformations must capture all closure relations of that module. The following construction assembles the module from pairwise path comparisons. Automatic extraction of such modules and path families from a general mechanism graph is outside the scope of this paper.

\subsection{Two-Path Closure Element}
\label{subsec:two-path-closure}
The closure condition of a closed kinematic chain is commonly expressed as an ordered product of relative homogeneous transformations around the kinematic loop, with the product equal to the identity~\cite{chen2005}. In the proposed framework, the same closure condition is instead formulated using two transformation products assembled separately along paths with common initial and terminal frames. The resulting two-path representation constitutes the elementary closure element of the framework. This reformulation is necessary because it permits the two paths to be assembled and differentiated separately before their path-wise differential contributions are combined through the closure mismatch.

For this purpose, consider a two-path closure element indexed by $\ell$ whose local generalized coordinate vector is extracted from the module-level coordinates using the selection matrix $\boldsymbol{\mathcal{P}}_{\!\!\ell}$, i.e.
\begin{equation}
   \widehat{\boldsymbol{q}}_m^{\,\lgroup \ell \rgroup}=\boldsymbol{\mathcal{P}}_{\!\!\ell} \widehat{\boldsymbol{q}}_m=\mathrm{col}\left(\widehat{\boldsymbol{q}}_{m,a}^{\,\lgroup \ell \rgroup} , \widehat{\boldsymbol{q}}_{m,p}^{\,\lgroup \ell \rgroup}\right).
   \label{eq:element-coordinates}
\end{equation}
Let $\boldsymbol{T}_\ell^{\langle + \rangle}\left(\widehat{\boldsymbol{q}}_m^{\,\lgroup \ell \rgroup}\right) \in \mathrm{SE}(3)$ and $\boldsymbol{T}_\ell^{\langle - \rangle}\left(\widehat{\boldsymbol{q}}_m^{\,\lgroup \ell \rgroup}\right) \in \mathrm{SE}(3)$ denote the ordered products of homogeneous transformation matrices along the two individual paths forming closure element $\ell$. The construction of the constituent homogeneous transformation matrices is described in Appendix~\ref{app:local-transformations}.

Equality of the terminal-frame poses obtained by traversing the two paths is imposed through the pairwise closure condition
\begin{equation}
    \left(\boldsymbol{T}_\ell^{\langle - \rangle}\left( \widehat{\boldsymbol{q}}_m^{\,\lgroup \ell \rgroup}\right)\right)^{-1}\boldsymbol{T}_\ell^{\langle + \rangle}\left( \widehat{\boldsymbol{q}}_m^{\,\lgroup \ell \rgroup}\right)=\boldsymbol{I}_4.
    \label{eq:closure-condition}
\end{equation}
Away from closure, the relative transformation on the left-hand side of \eqref{eq:closure-condition} differs from the identity and represents the closure mismatch. Accordingly, define
\begin{equation}
    \boldsymbol{\Delta}_\ell\left( \widehat{\boldsymbol{q}}_m^{\,\lgroup \ell \rgroup}\right):=\left(\boldsymbol{T}_\ell^{\langle - \rangle}\left( \widehat{\boldsymbol{q}}_m^{\,\lgroup \ell \rgroup}\right)\right)^{-1}\boldsymbol{T}_\ell^{\langle + \rangle}\left( \widehat{\boldsymbol{q}}_m^{\,\lgroup \ell \rgroup}\right).
    \label{eq:closure-mismatch}
\end{equation}
At closure, the two terminal-frame poses coincide, and the closure mismatch therefore equals the identity transformation. To represent the closure mismatch in local vector coordinates, define the closure residual as 
\begin{equation}
   \boldsymbol{\rho}_\ell\left( \widehat{\boldsymbol{q}}_m^{\,\lgroup \ell \rgroup}\right):=\operatorname{Log}\left(\boldsymbol{\Delta}_\ell\left( \widehat{\boldsymbol{q}}_m^{\,\lgroup \ell \rgroup}\right)\right)^\vee\in \mathbb{R}^6.
    \label{eq:closure-residual}
\end{equation}
Here, $\operatorname{Log}:\mathcal{U}\subset\mathrm{SE}(3)\rightarrow\mathfrak{se}(3)$ denotes the selected local branch of the Lie-group logarithm map, and $\left(\cdot\right)^\vee:\mathfrak{se}(3)\rightarrow\mathbb{R}^6$ denotes the vee map. Throughout this work, the vector representation of $\mathfrak{se}(3)$ follows the angular-first convention. Consequently, at closure
\begin{equation}
    \boldsymbol{\rho}_\ell\left( \widehat{\boldsymbol{q}}_m^{\,\lgroup \ell \rgroup}\right)=\operatorname{Log}\left(\boldsymbol{I}_4\right)^\vee=\boldsymbol{0}_6.
\end{equation}
Since the Lie-group logarithm is not globally single-valued, $\boldsymbol{\rho}_\ell$ provides a local coordinate representation of the closure mismatch and is valid only within the domain of the selected logarithm branch. In particular, under the principal-logarithm convention, the rotational component of $\boldsymbol{\rho}_\ell$ is uniquely represented for relative rotation angles strictly below $\pi$. At a relative rotation angle of $\pi$, the rotational logarithm is nonunique, and configurations near this branch boundary may require careful numerical handling~\cite{gebhardt2023,muller2025}.

Determining a locally independent set of closure constraints from the closure residual requires examining its differential with respect to the kinematic coordinates. The residual differential is therefore constructed next.

\subsection{Closure Element Differentials}
The closure residual represents the mismatch between the two paths in local vector coordinates, but it does not by itself describe how this mismatch changes under variations of the closure-element coordinates. Its differential is required to reveal the local rank of the closure relation and select independent residual coordinates, construct the Jacobians needed for passive-coordinate acquisition, and derive the active-to-passive differential map used to form the active-to-module-coordinate differential mapping.

For this purpose, the differential contributions of the two path transformations are first expressed at the group level. The inverse left Jacobian of $\mathrm{SE}(3)$ then maps the resulting group-level differential to the differential of the logarithmic closure residual. The resulting Path-Assembled Closure Differential Mapping (PACDM) relates the closure-element coordinate rates directly to the rate of change of the logarithmic closure residual.

Let $\sigma \in \left\{+,-\right\}$ index the two paths of closure element $\ell$. Each path transformation consists of $n_\ell^{\langle \sigma \rangle}$ local transformations $\boldsymbol{\mathcal{T}}_{\!\!\ell,k}^{\langle \sigma \rangle}\left(\widehat{\boldsymbol{q}}_m^{\,\lgroup \ell \rgroup}\right)\in \mathrm{SE}(3)$. Consequently, with the factors ordered according to the traversal of the path,
\begin{equation}
    \boldsymbol{T}_\ell^{\langle \sigma \rangle}\left(\widehat{\boldsymbol{q}}_m^{\,\lgroup \ell \rgroup}\right)=\prod_{k=1}^{n_\ell^{\langle \sigma \rangle}}\boldsymbol{\mathcal{T}}_{\!\!\ell,k}^{\langle \sigma \rangle}\left(\widehat{\boldsymbol{q}}_m^{\,\lgroup \ell \rgroup}\right),
    \label{eq:local-transforms}
\end{equation}
where each local transformation is either a rigid transformation, independent of all generalized coordinates, or a joint transformation, dependent on one or more coordinates of $\widehat{\boldsymbol{q}}_m^{\,\lgroup \ell \rgroup}$. Let $\widehat{q}_m^{\,\lgroup \ell\mid j \rgroup}$ denote the $j$-th coordinate of the closure element coordinate vector $\widehat{\boldsymbol{q}}_m^{\,\lgroup \ell \rgroup}$. We define the right-trivialized derivative of the $k$-th local transformation with respect to $\widehat{q}_m^{\,\lgroup \ell\mid j \rgroup}$ as
\begin{equation}
    \boldsymbol{\xi}_{\ell,k\mid j}^{\langle \sigma \rangle}:= \left[\dfrac{\partial \boldsymbol{\mathcal{T}}_{\!\!\ell,k}^{\langle \sigma \rangle}}{\partial \widehat{q}_m^{\,\lgroup \ell\mid j \rgroup}}\left(\boldsymbol{\mathcal{T}}_{\!\!\ell,k}^{\langle \sigma \rangle}\right)^{-1}\right]^{\large \,\vee}\in\mathbb{R}^6.
    \label{eq:right-trivialized-derivative}
\end{equation}

To transport the local right-trivialized derivatives to the common frame of the complete-path derivative, let $\boldsymbol{A}_{\ell,k}^{\langle \sigma \rangle}$ denote the prefix transformation preceding the $k$-th local transformation, defined as
\begin{equation}
    \boldsymbol{A}_{\ell,k}^{\langle \sigma \rangle}:=\prod_{i=1}^{k-1}\boldsymbol{\mathcal{T}}_{\!\!\ell,i}^{\langle \sigma \rangle}, \qquad \boldsymbol{A}_{\ell,1}^{\langle \sigma \rangle}=\boldsymbol{I}_4.
    \label{eq:prefix}
\end{equation}
These prefixes transport the local right-trivialized derivatives into the common frame associated with the complete path's right-trivialized derivative. Rigid transformations remain in these prefixes despite having zero local derivatives and therefore affect the transport of subsequent derivative contributions. As a result, the right-trivialized derivative of the complete path with respect to $\widehat{q}_m^{\,\lgroup \ell\mid j \rgroup}$ can be obtained as
\begin{equation}
    \boldsymbol{\eta}_{\ell, j}^{\langle \sigma \rangle}:=\left[\dfrac{\partial \boldsymbol{T}_\ell^{\langle \sigma \rangle}}{\partial \widehat{q}_m^{\,\lgroup \ell \mid j \rgroup}} \left(\boldsymbol{T}_\ell^{\langle \sigma \rangle} \right)^{-1} \right]^{\large\,\vee}=\sum_{i=1}^{n_\ell^{\langle \sigma \rangle}}\mathrm{Ad}_{\boldsymbol{A}_{\ell,i}^{\langle \sigma \rangle}} \boldsymbol{\xi}_{\ell,i\mid j}^{\langle \sigma \rangle}.
    \label{eq:complete-path-derivative}
\end{equation}
Here, $\mathrm{Ad}_{\boldsymbol{A}} \in \mathbb{R}^{6\times 6}$ denotes the adjoint matrix associated with $\boldsymbol{A}\in\mathrm{SE}(3)$ under the adopted vee convention. When $\widehat{q}_m^{\,\lgroup \ell \mid j \rgroup}$ appears in only one local transformation along path $\langle\sigma\rangle$, all other local derivative contributions vanish, and the summation in \eqref{eq:complete-path-derivative} reduces to the corresponding single adjoint-transformed term. If the coordinate does not appear along the path, then $\boldsymbol{\eta}_{\ell, j}^{\langle \sigma \rangle}=\boldsymbol{0}_6$.

With the complete-path right-trivialized derivatives obtained in \eqref{eq:complete-path-derivative}, the $j$-th column of the closure residual Jacobian is obtained as
\begin{equation}
    \dfrac{\partial \boldsymbol{\rho}_\ell}{\partial \widehat{q}_m^{\,\lgroup \ell \mid j \rgroup}}=\boldsymbol{J}_l^{-1}\left(\boldsymbol{\rho}_\ell\right) \left[\dfrac{\partial \boldsymbol{\Delta}_\ell}{\partial\widehat{q}_m^{\, \lgroup \ell \mid j \rgroup}} \boldsymbol{\Delta}_\ell^{-1}\right]^\vee=\boldsymbol{J}_l^{-1}\left(\boldsymbol{\rho}_\ell\right) \mathrm{Ad}_{\left(\boldsymbol{T}_\ell^{\langle - \rangle}\right)^{-1}} \left(\boldsymbol{\eta}_{\ell, j}^{\langle + \rangle}-\boldsymbol{\eta}_{\ell, j}^{\langle - \rangle} \right).
    \label{eq:pacdm-column}
\end{equation}
The inverse left Jacobian $\boldsymbol{J}_l^{-1}$ accounts for the differential of the nonlinear logarithm map, since a right-trivialized variation of the closure mismatch does not map directly to the residual variation. Related closed-form expressions for the differentials of the $\mathrm{SO}(3)$ and $\mathrm{SE}(3)$ exponential maps and their inverses are given in~\cite{muller2025} . To evaluate the inverse left Jacobian robustly, a norm-based switching strategy is adopted. When the small-adjoint matrix associated with the residual is sufficiently small, the inverse left Jacobian is evaluated using a truncated Bernoulli-series expansion. Away from this neighborhood, the left Jacobian is evaluated through a block matrix exponential and subsequently inverted through a linear solve. Specifically, let $\boldsymbol{Z}=\mathrm{ad}_{\boldsymbol{\rho}_\ell}$ and let $\varepsilon_J>0$ denote a prescribed series threshold. Here, $\mathrm{ad}_{\boldsymbol{\rho}_\ell}\in\mathbb{R}^{6\times6}$ denotes the small-adjoint matrix associated with $\boldsymbol{\rho}_\ell$. The Bernoulli-series evaluation \cite{iserles2000} is used when $\|\boldsymbol{Z}\|\leq\varepsilon_J$, whereas the block-exponential evaluation is used otherwise. For $\boldsymbol{\rho}_{\ell}\in\mathbb{R}^6$ satisfying $\|\operatorname{ad}_{\boldsymbol{\rho}_{\ell}}\|\leq\varepsilon_J$, the fourth-order Bernoulli-series approximation is
\begin{equation}
    \boldsymbol{J}_l^{-1}\left(\boldsymbol{\rho}_\ell\right)\approx \boldsymbol{I}_6-\dfrac{1}{2} \mathrm{ad}_{\boldsymbol{\rho}_\ell}+\dfrac{1}{12}\mathrm{ad}^2_{\boldsymbol{\rho}_\ell}-\dfrac{1}{720}\mathrm{ad}^4_{\boldsymbol{\rho}_\ell}.
    \label{eq:bernoulli-approximation}
\end{equation}
For residuals outside the series neighborhood, define the block matrix
\begin{equation}
    \boldsymbol{M}\left(\boldsymbol{\rho}_\ell\right):=\begin{bmatrix}
        \mathrm{ad}_{\boldsymbol{\rho}_\ell} & \boldsymbol{I}_6 \\[3pt] \boldsymbol{O}_6 & \boldsymbol{O}_6
    \end{bmatrix}.
    \label{eq:block-matrix}
\end{equation}
Here, $\boldsymbol{O}_6$ denotes the $6\times6$ zero matrix. The corresponding matrix exponential satisfies
\begin{equation}
    \exp\left(\boldsymbol{M}\left(\boldsymbol{\rho}_\ell\right)\right)=\begin{bmatrix}
        \exp\left(\mathrm{ad}_{\boldsymbol{\rho}_\ell}\right) & \boldsymbol{J}_l\!\left(\boldsymbol{\rho}_\ell\right) \\[5pt] \boldsymbol{O}_6 & \boldsymbol{I}_6
    \end{bmatrix}
    \label{eq:left-jacobian-block-exponential}
\end{equation}
from which $\boldsymbol{J}_l\left( \boldsymbol{\rho}_\ell\right)$ is extracted~\cite{vanloan1978}. The inverse left Jacobian is then obtained by solving
\begin{equation}
    \boldsymbol{J}_l\left(\boldsymbol{\rho}_\ell\right)
    \boldsymbol{X}_\ell
    =
    \boldsymbol{I}_6,
\end{equation}
for $\boldsymbol{X}_\ell
=
\boldsymbol{J}_l^{-1}\left(\boldsymbol{\rho}_\ell\right)$. Collecting the Jacobian columns obtained from \eqref{eq:pacdm-column} gives the PACDM Jacobian as
\begin{equation}
    \bar{\boldsymbol{R}}_m^{\lgroup \ell \rgroup}=\dfrac{\partial \boldsymbol{\rho}_\ell}{\partial \widehat{\boldsymbol{q}}_m^{\lgroup \ell \rgroup}}.
    \label{eq:pacdm-jacobian}
\end{equation}
The PACDM Jacobian characterizes the local sensitivity of the six-component closure residual to the closure-element coordinates. The raw residuals and their PACDM Jacobians are assembled at the module level before redundant residual coordinates are identified and removed.

\subsection{Multi-Path Module Construction}
\label{subsec:multi-path}

The preceding construction is formulated for an elementary pair of paths, but it extends directly to a module containing $N_m$ paths with common initial and terminal frames. Equality among the $N_m$ path transformations can be enforced using a graph-theoretically minimal set of $N_m-1$ two-path closure elements. A convenient construction selects one path as a reference and pairs it with each remaining path, thereby forming a star-shaped spanning tree on the set of paths. Each pair is processed separately through path assembly, closure-mismatch construction, and residual differentiation. The resulting raw residuals and Jacobians are then assembled at the module level, where a rank-revealing analysis identifies the independent residual coordinates.

Let $\ell_p$, with $p\in\left\{2,\ldots,N_m\right\}$, denote the closure element formed by pairing reference path $1$ with path $p$. Within closure element $\ell_p$, denote the transformation products along the reference path and the paired path by $\boldsymbol{T}_{\ell_p}^{\, \langle 1 \rangle}$ and $\boldsymbol{T}_{\ell_p}^{\, \langle p \rangle}$, respectively. The corresponding closure conditions are
\begin{equation}
    \left(\boldsymbol{T}_{\ell_p}^{\langle p \rangle}\left( \widehat{\boldsymbol{q}}_m^{\,\lgroup \ell_p \rgroup}\right)\right)^{-1}\boldsymbol{T}_{\ell_p}^{\langle 1 \rangle}\left( \widehat{\boldsymbol{q}}_m^{\,\lgroup \ell_p \rgroup}\right)=\boldsymbol{I}_4, \qquad p=2,\ldots,N_m.
    \label{eq:multi-path-closure-condition}
\end{equation}

For each closure element $\ell_p$, the closure mismatch, raw closure residual, and element-level PACDM Jacobian are constructed according to the preceding two-path formulation. The raw closure-element residuals are then stacked to form the module-level residual vector
\begin{equation}
    \bar{\boldsymbol{\rho}}_m
    =
    \mathrm{col}\left(
    \boldsymbol{\rho}_{\ell_2},
    \ldots,
    \boldsymbol{\rho}_{\ell_{N_m}}
    \right)
    \in
    \mathbb{R}^{6(N_m-1)}.
    \label{eq:raw-module-residual}
\end{equation}
The arguments of the element residuals are omitted here for compactness. Since the closure-element coordinates are extracted from the module-level coordinate vector according to
\begin{equation}
    \widehat{\boldsymbol{q}}_m^{\,\lgroup \ell_p \rgroup}
    =
    \boldsymbol{\mathcal{P}}_{\!\!\ell_p}
    \widehat{\boldsymbol{q}}_m,
\end{equation}
the corresponding element-level PACDM Jacobian is mapped to the module-coordinate space through
\begin{equation}
    \dfrac{\partial \boldsymbol{\rho}_{\ell_p}}
    {\partial \widehat{\boldsymbol{q}}_m}
    =
    \bar{\boldsymbol{R}}_m^{\lgroup \ell_p \rgroup}
    \boldsymbol{\mathcal{P}}_{\!\!\ell_p}.
    \label{eq:element-to-module-jacobian}
\end{equation}
The raw module-level PACDM Jacobian is therefore obtained by stacking the mapped element-level Jacobians as
\begin{equation}
    \bar{\boldsymbol{R}}_m
    =
    \dfrac{\partial \bar{\boldsymbol{\rho}}_m}
    {\partial \widehat{\boldsymbol{q}}_m}
    =
    \mathrm{col}\left(
    \bar{\boldsymbol{R}}_m^{\lgroup \ell_2 \rgroup}
    \boldsymbol{\mathcal{P}}_{\!\!\ell_2},
    \ldots,
    \bar{\boldsymbol{R}}_m^{\lgroup \ell_{N_m} \rgroup}
    \boldsymbol{\mathcal{P}}_{\!\!\ell_{N_m}}
    \right).
    \label{eq:raw-module-pacdm-jacobian}
\end{equation}

The raw module-level PACDM Jacobian characterizes the local sensitivity of all $6(N_m-1)$ residual components to the module coordinates. Its rank determines the local closure rank of the module, while a set of linearly independent rows identifies the residual coordinates to be retained. Although the spanning-tree construction provides a graph-theoretically minimal set of pairwise path-equality relations, the stacked logarithmic residual need not impose $6(N_m-1)$ independent constraints. Dependencies may occur both among the residual components of an individual closure element and among the residuals of different closure elements after their Jacobians are expressed in the common module-coordinate space. The number and choice of locally independent residual coordinates therefore depend on the geometry of the mechanism and on the considered configuration.

The raw module-level PACDM Jacobian is consequently subjected to a rank-revealing analysis. The rank of the full Jacobian and the rank of its passive-coordinate block serve distinct purposes. The former determines the local number of independent closure constraints, whereas the latter determines whether the selected passive coordinates can be locally resolved as functions of the active coordinates.

For this purpose, define the local closure rank of module $m$ at the considered configuration as
\begin{equation}
    r_m
    :=
    \mathrm{rank}\left(
    \bar{\boldsymbol{R}}_m
    \right).
    \label{eq:module-closure-rank}
\end{equation}
In numerical evaluation, $r_m$ is interpreted as the numerical rank determined under a prescribed tolerance. Partition the raw module-level PACDM Jacobian according to the active and passive module coordinates as
\begin{equation}
    \bar{\boldsymbol{R}}_m
    =
    \begin{bmatrix}
        \bar{\boldsymbol{R}}_{m,a}
        &
        \bar{\boldsymbol{R}}_{m,p}
    \end{bmatrix},
\end{equation}
where
\begin{equation}
    \bar{\boldsymbol{R}}_{m,a}
    =
    \dfrac{\partial\bar{\boldsymbol{\rho}}_m}
    {\partial\widehat{\boldsymbol{q}}_{m,a}},
    \qquad
    \bar{\boldsymbol{R}}_{m,p}
    =
    \dfrac{\partial\bar{\boldsymbol{\rho}}_m}
    {\partial\widehat{\boldsymbol{q}}_{m,p}}.
\end{equation}
Let $n_{m,p}$ denote the number of passive coordinates of module $m$. For the adopted active--passive partition to admit a locally unique passive-coordinate map, the physical configuration must satisfy
\begin{equation}
    r_m=n_{m,p},
    \qquad
    \mathrm{rank}\left(\bar{\boldsymbol{R}}_{m,p}\right)=n_{m,p}.
\end{equation}
Under these conditions, a set of $n_{m,p}$ residual coordinates is selected from the passive-coordinate block. Let
$\boldsymbol{S}_m\in\left\{0,1\right\}^{n_{m,p}\times6(N_m-1)}$
denote the corresponding row-selection matrix, obtained numerically from a pivoted rank-revealing factorization of $\bar{\boldsymbol{R}}_{m,p}^{T}$~\cite{golub2013}. The selected module-level residual is then
\begin{equation}
    \boldsymbol{c}_m
    =
    \boldsymbol{S}_m
    \bar{\boldsymbol{\rho}}_m
    \in
    \mathbb{R}^{n_{m,p}},
    \label{eq:selected-module-residual}
\end{equation}
and its PACDM Jacobian is
\begin{equation}
    \boldsymbol{R}_m
    =
    \boldsymbol{S}_m
    \bar{\boldsymbol{R}}_m
    =
    \begin{bmatrix}
        \boldsymbol{R}_{m,a}
        &
        \boldsymbol{R}_{m,p}
    \end{bmatrix}
    =
    \dfrac{\partial\boldsymbol{c}_m}
    {\partial\widehat{\boldsymbol{q}}_m},
    \label{eq:selected-module-pacdm-jacobian}
\end{equation}
where
\begin{equation}
    \boldsymbol{R}_{m,p}
    =
    \boldsymbol{S}_m\bar{\boldsymbol{R}}_{m,p}
\end{equation}
is square and nonsingular. For the local analysis about a regular physical configuration, the selected equations are assumed to be locally equivalent to the full raw closure equations. A sufficient condition is that the full closure Jacobian retains the constant rank \(n_{m,p}\) in a neighborhood of the configuration. Under this assumption, the selected nonsingular passive block defines the same local physical closure branch and permits the active-to-passive differential mapping to be formed. If the required physical rank conditions are not satisfied, or if the selected passive block is numerically singular under the adopted conditioning criterion, the declared active--passive partition is treated as singular at that configuration and the mapping is not formed.

The selection above concerns the recovered physical closure configuration. The residual equations used to acquire that configuration from an approximate passive-coordinate estimate are selected separately from the defected passive Jacobian at the artificial starting point, as described in the following section. For $N_m=2$, the module-level construction reduces directly to the elementary two-path closure formulation.

\section{Passive-Coordinate Resolution via Defect Homotopy}
\label{sec:defect-homotopy}

In mechanisms and robotic systems, the active coordinate values are typically available from sensor measurements. A rough initial estimate of the passive coordinate values can also be obtained relatively easily using practical information about the mechanism, such as passive-joint limits and visual inspection of the multibody-system configuration. Such information additionally helps identify the relevant passive-coordinate solution branch, since the nonlinear closure residual equations may admit multiple passive-coordinate solutions. With the measured active-coordinate and estimated passive-coordinate values, each two-path closure element generally produces a nonzero closure residual. Our approach to finding closure-consistent passive-coordinate values is to incorporate this initial closure mismatch as a \textit{defect}, thereby constructing a defected closure problem that is satisfied at the initial estimate. Using homotopy continuation, this defect is then gradually removed until the original closure problem is recovered~\cite{allgower1990}. The construction and solution of this defect-homotopy problem are developed in this section.

\subsection{Defect Homotopy}
Let $\widehat{\boldsymbol{q}}^0_m$ denote the selected initial module configuration, with its active-coordinate values held fixed during the passive-coordinate resolution. For every two-path closure element in this module, the initial closure mismatch is obtained using \eqref{eq:closure-mismatch} as
\begin{equation}
    \boldsymbol{\Delta}_\ell^0=\boldsymbol{\Delta}_\ell \left(\widehat{\boldsymbol{q}}_m^0\right).
\end{equation}
The corresponding initial defect vector is then defined as
\begin{equation}
    \boldsymbol{d}_\ell^0:=\mathrm{Log}\left(\boldsymbol{\Delta}_\ell^0\right)^\vee.
\end{equation}
The initial closure mismatch is assumed to lie in the domain of the selected logarithm branch, so that the corresponding exponential-logarithm composition recovers $\boldsymbol{\Delta}_\ell^0$. The defect transformation corresponding to the closure element is defined as
\begin{equation}
\boldsymbol{\mathcal{D}}_\ell\left(\lambda\right):=\mathrm{Exp}\left[\left(1-\lambda\right)\boldsymbol{d}_\ell^0\right],   
\label{eq:defect-map}
\end{equation}
where $\lambda \in \left[0,1\right]$ denotes the homotopy parameter. It follows that
\begin{equation}
    \boldsymbol{\mathcal{D}}_\ell\left(0\right)=\boldsymbol{\Delta}_\ell^0, \qquad \boldsymbol{\mathcal{D}}_\ell\left(1\right)=\boldsymbol{I}_4,
    \label{eq:defect-transform-interpretation}
\end{equation}
so that the defect transformation interpolates from the initial closure mismatch to the identity transformation. Consequently, the defected closure mismatch becomes
\begin{equation}
    \boldsymbol{\Lambda}_\ell\left(\widehat{\boldsymbol{q}}_m,\lambda\right)=\left[\boldsymbol{T}_\ell^{\langle - \rangle}\left(\widehat{\boldsymbol{q}}_m\right)\boldsymbol{\mathcal{D}}_\ell\left(\lambda\right)\right]^{-1}\boldsymbol{T}_\ell^{\langle + \rangle}\left(\widehat{\boldsymbol{q}}_m\right).
    \label{eq:defected-mismatch}
\end{equation}
At the initial module configuration, \eqref{eq:defect-transform-interpretation} gives
\begin{equation}
    \boldsymbol{\Lambda}_\ell\left(\widehat{\boldsymbol{q}}_m^0,0\right)=\boldsymbol{I}_4,
\end{equation}
meaning that the selected initial module configuration is an exact solution of the defected closure problem at $\lambda=0$, whereas
\begin{equation}
    \boldsymbol{\Lambda}_\ell\left(\widehat{\boldsymbol{q}}_m,1\right)=\boldsymbol{\Delta}_\ell\left(\widehat{\boldsymbol{q}}_m\right),
\end{equation}
showing that, at $\lambda=1$, the artificial defect vanishes and the original closure problem is recovered. The defect residual corresponding to \eqref{eq:defected-mismatch} is therefore
\begin{equation}
    \boldsymbol{\varrho}_\ell\left(\widehat{\boldsymbol{q}}_m,\lambda\right):=\mathrm{Log}\left[\boldsymbol{\Lambda}_\ell\left(\widehat{\boldsymbol{q}}_m,\lambda\right)\right]^\vee.
    \label{eq:defected-residual}
\end{equation}
The homotopy therefore does not alter the mechanism itself, but deforms the closure equations used for passive-coordinate resolution. The continuous variation of $\lambda$ transforms an artificial problem with a known solution at $\lambda=0$ into the physical closure problem at $\lambda=1$.

Expanding \eqref{eq:defected-mismatch} yields
\begin{equation}
    \boldsymbol{\Lambda}_\ell\left(\widehat{\boldsymbol{q}}_m,\lambda\right)=\boldsymbol{\mathcal{D}}_\ell^{-1}\left(\lambda\right)\left[\boldsymbol{T}_\ell^{\langle - \rangle}\left(\widehat{\boldsymbol{q}}_m\right)\right]^{-1}\boldsymbol{T}_\ell^{\langle+\rangle}\left(\widehat{\boldsymbol{q}}_m\right)=\boldsymbol{\mathcal{D}}_\ell^{-1}\left(\lambda\right)\boldsymbol{\Delta}_\ell\left(\widehat{\boldsymbol{q}}_m\right).
    \label{eq:defect-expanded-form}
\end{equation}
Since $\boldsymbol{\mathcal{D}}_\ell\left(\lambda\right)$ is fixed with respect to the coordinates at each fixed $\lambda$, its introduction does not contribute an additional coordinate-derivative term. As a result, the right-trivialized derivative of each path in the closure element remains unchanged, and the PACDM relations can therefore be extended directly to the defected closure construction. Since the defect transformation is appended to path $\langle -\rangle$ in the defected closure construction, the corresponding adjoint transport is altered, and the defected PACDM columns from \eqref{eq:pacdm-column} become
\begin{equation}
            \dfrac{\partial \boldsymbol{\varrho}_\ell}{\partial \widehat{q}_m^{\,\lgroup \ell \mid j \rgroup}}=\boldsymbol{J}_l^{-1}\left(\boldsymbol{\varrho}_\ell\right) \mathrm{Ad}_{\left(\boldsymbol{T}_\ell^{\langle - \rangle}\boldsymbol{\mathcal{D}}_\ell\right)^{-1}} \left(\boldsymbol{\eta}_{\ell, j}^{\langle + \rangle}-\boldsymbol{\eta}_{\ell, j}^{\langle - \rangle} \right).
    \label{eq:defected-pacdm-column}
\end{equation}
Concatenating these columns yields the defected PACDM Jacobian 
\begin{equation}
    \bar{\boldsymbol{\mathcal{R}}}_m^{\lgroup \ell \rgroup}:=\dfrac{\partial \boldsymbol{\varrho}_\ell}{\partial \widehat{\boldsymbol{q}}_m^{\lgroup \ell \rgroup}},
    \label{eq:defected-pacdm}
\end{equation}
for the closure element $\ell$ in module $m$. The closure-element defected residuals are stacked to form the module-level defected residual $\bar{\boldsymbol{\varrho}}_m$, while the corresponding defected PACDM Jacobians are mapped to the common module-coordinate space and stacked to form $\bar{\boldsymbol{\mathcal{R}}}_m$, following the procedure described in Subsection \ref{subsec:multi-path}.

\subsection{Passive-Coordinate Corrector and Continuation}

At $\lambda=0$, the defect construction makes the initial configuration
$\widehat{\boldsymbol{q}}_m^0$ an exact solution of the artificial closure problem. Let
$n_{m,p}=\dim\left(\widehat{\boldsymbol{q}}_{m,p}\right)$ denote the number of passive coordinates of module $m$. The residual coordinates used for passive-coordinate acquisition are selected from the defected passive Jacobian evaluated at this exact artificial starting point. Define
\begin{equation}
    \bar{\boldsymbol{\mathcal{R}}}_{m,p}^{\,0}
    :=
    \left.
    \dfrac{\partial\bar{\boldsymbol{\varrho}}_m}
    {\partial\widehat{\boldsymbol{q}}_{m,p}}
    \right|_{\left(\widehat{\boldsymbol{q}}_m^0,\,0\right)}.
\end{equation}
A regular starting branch requires
\begin{equation}
    \mathrm{rank}\left(
        \bar{\boldsymbol{\mathcal{R}}}_{m,p}^{\,0}
    \right)
    =
    n_{m,p}.
\end{equation}
A row-selection matrix
$\boldsymbol{S}_m^{\,0}\in
\left\{0,1\right\}^{n_{m,p}\times6(N_m-1)}$
is obtained from a pivoted rank-revealing factorization of
$\left(\bar{\boldsymbol{\mathcal{R}}}_{m,p}^{\,0}\right)^T$
such that the selected passive block is nonsingular under the prescribed numerical conditioning criterion. The superscript $0$ distinguishes this acquisition selector from the physical row-selection matrix $\boldsymbol{S}_m$ defined in Subsection \ref{subsec:multi-path}. The intended passive-coordinate branch is identified by the selected initial passive-coordinate estimate together with the prescribed admissible passive-coordinate intervals, and the continuation seeks the physical solution connected continuously to this artificial starting configuration.

The selected row set is held fixed throughout the corresponding continuation from $\lambda=0$ to $\lambda=1$. Accordingly, the selected module-level defect residual is defined as
\begin{equation}
    \boldsymbol{\varrho}_m
    \left(
        \widehat{\boldsymbol{q}}_m,\lambda
    \right)
    :=
    \boldsymbol{S}_m^{\,0}
    \bar{\boldsymbol{\varrho}}_m
    \left(
        \widehat{\boldsymbol{q}}_m,\lambda
    \right),
\end{equation}
with the corresponding selected defected passive Jacobian
\begin{equation}
    \boldsymbol{\mathcal{R}}_{m,p}
    \left(
        \widehat{\boldsymbol{q}}_m,\lambda
    \right)
    :=
    \boldsymbol{S}_m^{\,0}
    \bar{\boldsymbol{\mathcal{R}}}_{m,p}
    \left(
        \widehat{\boldsymbol{q}}_m,\lambda
    \right)
    =
    \dfrac{\partial\boldsymbol{\varrho}_m}
    {\partial\widehat{\boldsymbol{q}}_{m,p}}.
\end{equation}

For clarity, throughout this subsection the defect residual is regarded as a function of the passive coordinates and the homotopy parameter, with the active-coordinate values fixed. At a fixed value of $\lambda$, introducing damping into the linearized passive-coordinate correction gives
\begin{equation}
    \delta\widehat{\boldsymbol{q}}_{m,p}
    =
    \operatorname*{arg\,min}_{\delta\widehat{\boldsymbol{q}}_{m,p}}
    \left[
        \dfrac{1}{2}
        \left\|
            \boldsymbol{\varrho}_m
            \left(
                \widehat{\boldsymbol{q}}_{m,p},\lambda
            \right)
            +
            \boldsymbol{\mathcal{R}}_{m,p}
            \delta\widehat{\boldsymbol{q}}_{m,p}
        \right\|_2^2
        +
        \dfrac{\mu}{2}
        \left\|
            \delta\widehat{\boldsymbol{q}}_{m,p}
        \right\|_2^2
    \right],
\end{equation}
which is the damped Gauss--Newton problem with damping parameter $\mu>0$~\cite{nocedal2006}. The first-order optimality condition yields the damped normal equations
\begin{equation}
    \left[
        \boldsymbol{\mathcal{R}}_{m,p}^{T}
        \boldsymbol{\mathcal{R}}_{m,p}
        +
        \mu\boldsymbol{I}_{n_{m,p}}
    \right]
    \delta\widehat{\boldsymbol{q}}_{m,p}
    =
    -
    \boldsymbol{\mathcal{R}}_{m,p}^{T}
    \boldsymbol{\varrho}_m
    \left(
        \widehat{\boldsymbol{q}}_{m,p},\lambda
    \right).
    \label{eq:gauss-newton-corrector}
\end{equation}
Only the passive coordinates are corrected, while the prescribed active-coordinate values remain fixed.

The unweighted selected residual and correction equations above assume appropriate numerical scaling of both residual and coordinate components. Where substantially different characteristic scales occur, the corresponding residual and coordinate variables may be nondimensionalized before applying the corrector. The solution of \eqref{eq:gauss-newton-corrector} provides a correction to the current passive-coordinate estimate.

The passive-coordinate estimate is iteratively updated using these corrections at the fixed homotopy parameter. After each update, the selected defect residual and its passive-coordinate Jacobian are reevaluated, and the correction is repeated until prescribed convergence criteria are satisfied. Practical restrictions, such as admissible passive-coordinate intervals, may be imposed during the numerical correction to restrict the iteration to the admissible region associated with the selected solution branch. The particular bound-handling, damping-update, and convergence strategies are numerical implementation choices and do not alter the defect-homotopy formulation.

Upon convergence at the fixed homotopy parameter, the resulting coordinates define the passive-coordinate solution
$\widehat{\boldsymbol{q}}_{m,p}^{*}(\lambda)$. The acquisition row-selection matrix $\boldsymbol{S}_m^{\,0}$ remains fixed during the continuation associated with the selected initial configuration, while the selected defect residual and selected passive-coordinate Jacobian are reevaluated at each corrector iteration. The converged passive coordinates are then returned to the continuation procedure and provide the starting configuration for advancing the homotopy parameter.

The continuation procedure wraps the fixed-$\lambda$ corrector and advances the homotopy parameter from $\lambda=0$ to $\lambda=1$. After successful correction at a continuation point, the converged passive coordinates are reused without extrapolation as the initial estimate for the next homotopy value, corresponding to a zero-order predictor. Consequently, no explicit derivative of the defected residual with respect to $\lambda$ is required. The fixed-$\lambda$ corrector is considered successful when the selected defect residual satisfies the prescribed convergence tolerance. Stagnation of the passive-coordinate update while this residual remains above tolerance, or exhaustion of the maximum number of corrector iterations, is treated as corrector failure.

At each trial homotopy value, convergence of the selected corrector is followed by two additional acceptance checks. The complete raw defect residual must satisfy
\begin{equation}
    \left\|
        \bar{\boldsymbol{\varrho}}_m
        \left(
            \widehat{\boldsymbol{q}}_m,\lambda
        \right)
    \right\|_{\infty}
    \leq
    \varepsilon_{\mathrm{raw}},
\end{equation}
and the selected defected passive Jacobian must satisfy the prescribed reciprocal-condition criterion. A continuation point is accepted only when the selected corrector has converged and both conditions are satisfied. Otherwise, the trial point is rejected, the homotopy increment is reduced, and the step is retried. If a rejected trial point would require a continuation increment below the prescribed minimum, the homotopy run is terminated and passive-coordinate acquisition is declared unsuccessful for the selected initial estimate and admissible bounds. The complete raw-residual check provides a numerical consistency test that the fixed selected residual coordinates continue to represent the complete defected closure equations along the accepted continuation path.

Passive-coordinate bounds are enforced directly during each trial correction, whereas admissibility of the adopted logarithm branch is retained as a regularity assumption of the continuation. Successful continuation assumes that the intended solution branch can be represented continuously over the traced interval, remains within the prescribed passive-coordinate bounds and the adopted logarithm domain, and retains a nonsingular selected defected passive Jacobian along the branch. The fixed acquisition selector $\boldsymbol{S}_m^{0}$ is retained throughout a homotopy run. Reduction of the homotopy increment improves the numerical resolution of difficult portions of the continuation path, but does not provide a global guarantee of continuation through a loss of regularity or a branch boundary.

At $\lambda=1$, the artificial defect vanishes and the original physical closure problem is recovered. Satisfaction of the complete raw physical closure residual $\bar{\boldsymbol{\rho}}_m$ is then checked independently of the selected corrector residual. After successful closure recovery, the physical PACDM Jacobian is reevaluated and the residual rows used for the active-to-passive differential mapping are selected anew from its passive-coordinate block according to the procedure described in Subsection \ref{subsec:multi-path}.

\section{Kinematic Reduction and Active-to-Passive Mapping}
\label{sec:active-to-passive}

Recovering a closure-consistent physical configuration for each closed-chain module enables the subsequent kinematic reduction. At the physical configuration, the artificial defect is no longer present, and the original nondefected closure residual and PACDM Jacobian are reevaluated. Their local rank and independent constraint structure provide the regularity conditions required to express the passive-coordinate differentials in terms of the active-coordinate differentials and thereby construct the reduced kinematics. 

\subsection{Kinematic Reduction}
After the defect-homotopy continuation reaches $\lambda=1$ and the fixed-$\lambda$ corrector converges, the closure-consistent physical configuration of module $m$,
\begin{equation}
    \widehat{\boldsymbol{q}}_m^*=\operatorname{col}\left(\widehat{\boldsymbol{q}}_{m,a},\widehat{\boldsymbol{q}}_{m,p}^*\right),
\end{equation}
is recovered. The original nondefected closure residual is first reevaluated at $\widehat{\boldsymbol{q}}_m^*$ to verify that the physical closure condition is satisfied to the prescribed tolerance. The corresponding nondefected raw module-level PACDM Jacobian is then evaluated from \eqref{eq:raw-module-pacdm-jacobian} and partitioned into active-coordinate and passive-coordinate columns as
\begin{equation}
    \bar{\boldsymbol{R}}_m=\left[\bar{\boldsymbol{R}}_{m,a}\quad\bar{\boldsymbol{R}}_{m,p}\right].
\end{equation}
Because the local closure rank and the independent residual-coordinate set depend on the mechanism configuration, the rank-revealing analysis introduced in Subsection \ref{subsec:multi-path} is reevaluated at $\widehat{\boldsymbol{q}}_m^*$. The resulting numerical rank $r_m$ therefore gives the number of locally independent scalar closure constraints at the recovered physical configuration.

Let $n_{m,p}=\dim\left(\widehat{\boldsymbol{q}}_{m,p}\right)$ denote the number of passive coordinates in the module $m$. For the module configuration to admit the intended local parametrization by the active coordinates, the closure rank must satisfy
\begin{equation}
    r_m=n_{m,p},
\end{equation}
and the passive-coordinate block of the raw PACDM Jacobian must have full column rank,
\begin{equation}
    \operatorname{rank}\left(\bar{\boldsymbol{R}}_{m,p}\right)=n_{m,p}.
\end{equation}
Together, these conditions ensure that the number of independent closure constraints matches the number of passive coordinates and that the constraints provide independent sensitivity with respect to all passive-coordinate directions. Loss of full column rank in $\bar{\boldsymbol{R}}_{m,p}$ invalidates the intended local resolution of the passive coordinates with respect to the active coordinates. If the overall closure rank simultaneously falls below its regular value $n_{m,p}$, the configuration is singular; otherwise, the loss reflects an unsuitable active/passive coordinate partition at that configuration. Applying the row-selection matrix $\boldsymbol{S}_m$ then gives the selected module-level PACDM Jacobian, which retains the same active-passive column partition,
\begin{equation}
    \boldsymbol{R}_m=\boldsymbol{S}_m\bar{\boldsymbol{R}}_m=\left[\boldsymbol{R}_{m,a}\quad\boldsymbol{R}_{m,p}\right]. 
\end{equation}
Because $\bar{\boldsymbol{R}}_{m,p}$ has full column rank, the row-selection matrix $\boldsymbol{S}_m$ can be chosen such that the selected passive block $\boldsymbol{R}_{m,p}=\boldsymbol{S}_m \bar{\boldsymbol{R}}_{m,p}$ is square and nonsingular. 

\subsection{Active-to-Passive Coordinate Mapping}
At a regular physical configuration, the selected module-level closure condition can be written as
\begin{equation}
    \boldsymbol{c}_m\left(\widehat{\boldsymbol{q}}_{m,a},\widehat{\boldsymbol{q}}_{m,p}\right)=\boldsymbol{0}.
\end{equation}
Because the selected passive-coordinate Jacobian $\boldsymbol{R}_{m,p}$ is nonsingular, the implicit function theorem guarantees, locally on the selected solution branch, the existence of a passive-coordinate map
\begin{equation}
    \widehat{\boldsymbol{q}}_{m,p}=\boldsymbol{\psi}_m\left(\widehat{\boldsymbol{q}}_{m,a}\right).
\end{equation}
Differentiating the selected module-level closure condition gives
\begin{equation}
    \boldsymbol{R}_{m,a} d\widehat{\boldsymbol{q}}_{m,a} + \boldsymbol{R}_{m,p} d\widehat{\boldsymbol{q}}_{m,p}=\boldsymbol{0}.
\end{equation}
Along the local solution branch, $d\widehat{\boldsymbol{q}}_{m,p}=\boldsymbol{D}\boldsymbol{\psi}_m d\widehat{\boldsymbol{q}}_{m,a}$. Substitution into the closure differential therefore gives
\begin{equation}
    \boldsymbol{D}\boldsymbol{\psi}_m=-\boldsymbol{R}_{m,p}^{-1} \boldsymbol{R}_{m,a}.
\end{equation}
The PACDM blocks in this expression are evaluated on the closure-consistent local solution branch. The corresponding intermediate Jacobian, which maps active-coordinate variations to the complete module-coordinate variations, is therefore
\begin{equation}
    \boldsymbol{E}_m=\dfrac{\partial \widehat{\boldsymbol{q}}_m}{\partial \widehat{\boldsymbol{q}}_{m,a}}=\begin{bmatrix}
        \boldsymbol{I}_{n_{m,a}}\\ \boldsymbol{D}\boldsymbol{\psi}_m
    \end{bmatrix}.
\end{equation}
Here, $n_{m,a}$ denotes the number of active coordinates in module $m$. The defect-homotopy procedure provides branch acquisition from a general passive-coordinate estimate. Once a physical configuration on the desired branch has been obtained, the local differential mapping $\boldsymbol{D}\boldsymbol{\psi}_m$ can be used to form a first-order prediction of the passive-coordinate changes associated with small increments of the active coordinates. Accordingly, for the $k$-th step of a trajectory, the passive coordinates may be predicted from the preceding closure-consistent configuration as
\begin{equation}
\widehat{\boldsymbol{q}}_{m,p,\mathrm{pred}}^{(k)}
=
\widehat{\boldsymbol{q}}_{m,p}^{(k-1)}
+
\boldsymbol{D}\boldsymbol{\psi}_m^{(k-1)}
\left(
\widehat{\boldsymbol{q}}_{m,a}^{(k)}
-
\widehat{\boldsymbol{q}}_{m,a}^{(k-1)}
\right).
\label{eq:continuation-predictor}
\end{equation}
This predictor is the first-order linearization of the local passive-coordinate map about the preceding closure-consistent configuration; the subsequent physical corrector restores the nonlinear closure equations at the new trajectory point. Because this relation constitutes a first-order prediction for a finite active-coordinate increment, the predicted passive coordinates are subsequently used as the initial estimate for a direct correction on the physical closure manifold by solving
\begin{equation}
\boldsymbol{c}_m
\left(
\widehat{\boldsymbol{q}}_{m,a}^{(k)},
\widehat{\boldsymbol{q}}_{m,p}^{(k)}
\right)
=
\boldsymbol{0}.
\label{eq:continuation-corrector}
\end{equation}
At trajectory step $k>1$, the physical row-selection matrix $\boldsymbol{S}_m^{(k-1)}$ obtained at the preceding closure-consistent configuration is retained throughout the direct correction of the predicted configuration. After convergence and acceptance of the configuration at step $k$, the raw physical PACDM Jacobian is reevaluated, its physical rank and passive-block rank are checked, and a new physical row-selection matrix $\boldsymbol{S}_m^{(k)}$ is obtained from the passive-coordinate block. The resulting selector and differential mapping are then retained for prediction and correction at the subsequent trajectory step.

A direct trajectory-corrector solution is accepted only if the selected physical residual has converged, the complete raw physical closure residual satisfies the prescribed tolerance, and the passive block associated with the retained physical selector satisfies the prescribed reciprocal-condition criterion. If any of these conditions is not satisfied, the direct correction is treated as unsuccessful and passive-coordinate branch reacquisition through defect homotopy is invoked.

Thus, nearby configurations along continuous motion need not be reacquired through a complete defect-homotopy continuation at every step. Instead, the differential mapping serves as a local first-order predictor, followed by direct correction onto the physical closure conditions. After successful correction, the PACDM blocks and $D\boldsymbol{\psi}_m$ are reevaluated at the new closure-consistent configuration for use at the subsequent step.

Accordingly, the passive-coordinate dependence is incorporated into the module kinematics without retaining the passive coordinates as independent variables. Applying this construction to the closed-chain modules completes the proposed kinematic reduction and provides the quantities required for expressing the mechanism kinematics in terms of its active coordinates. The following section evaluates the proposed formulation numerically by examining the recovery of closure-consistent configurations, the associated rank conditions, the resulting reduced kinematic mappings, and their continuation along prescribed motion.

\section{Simulation and Results}
\label{sec:simulation}
In this section, we apply the proposed framework to a 7-degree-of-freedom (7-DoF) robotic manipulator that contains two closed-chain modules. One module consists of two paths and the other consists of three paths. The numerical study is conducted in two parts. The first part examines passive-coordinate acquisition through defect homotopy in different fixed configurations. The second part explores continuation using the local first-order approximation following the initial passive-coordinate acquisition via defect homotopy. 

\begin{figure}[pos=htbp]
    \centering
    \includegraphics[width=\linewidth,trim={0cm 0cm 0cm 0cm},clip]{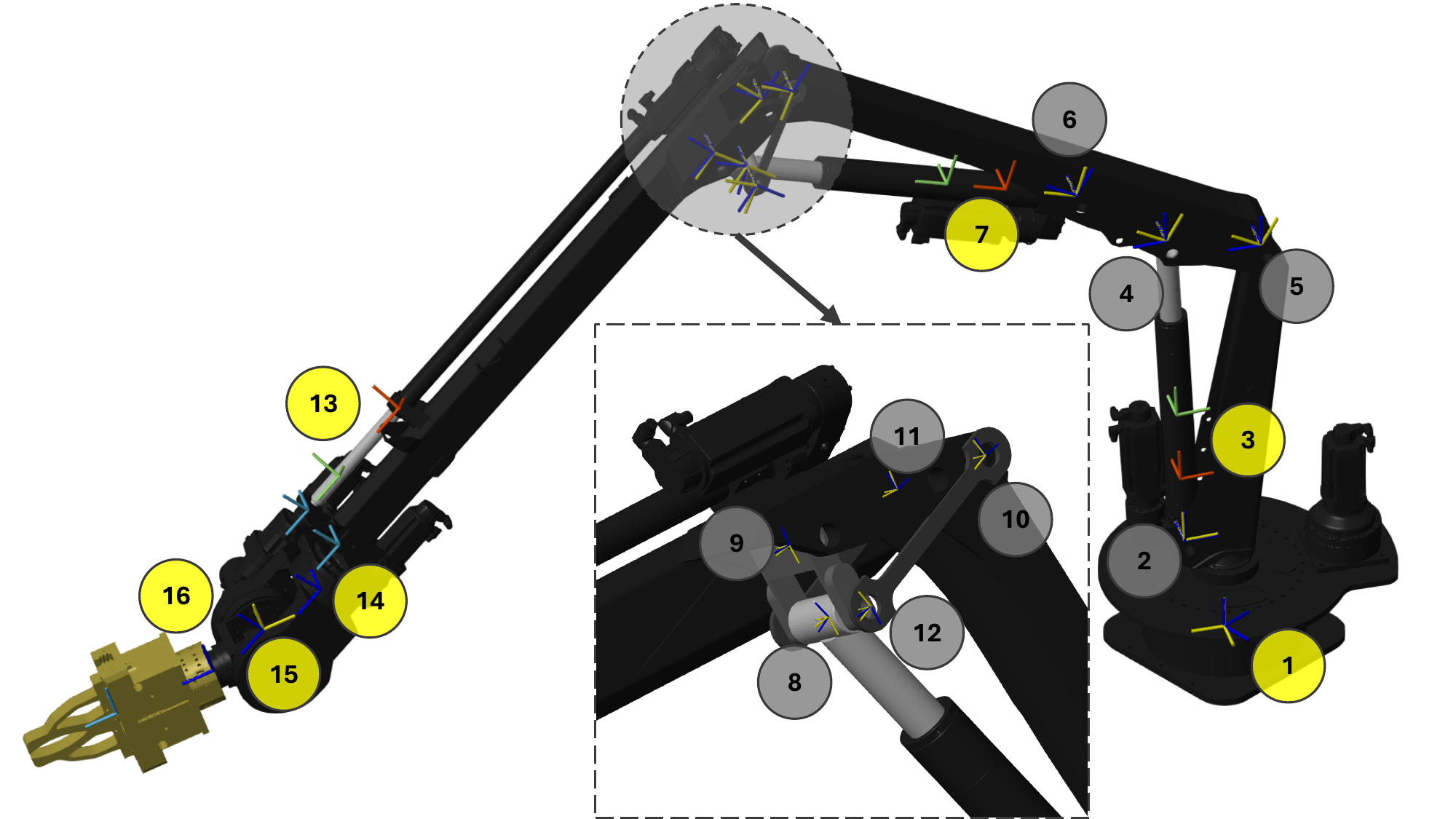}
    \caption{Simscape Multibody model of the HIAB-046 7-DoF heavy-duty robotic manipulator with the joint and frame annotations used in the numerical validation. A close-up view of the four-bar linkage is provided for clarity. The physical parameters of the robot are omitted due to a non-disclosure agreement (NDA).}
    \label{fig:simscape-robot-model}
\end{figure}

The 3D model of the robot in the Simscape Multibody environment is shown in Fig. \ref{fig:simscape-robot-model}. The active and passive revolute and prismatic joints are marked with a yellow and a gray circle around the joint number, respectively. The base and follower frames of the revolute joints are marked in blue and yellow, respectively. The base and follower frames of the prismatic joints are marked in orange and green, respectively. The auxiliary frames that are not attached to any joint are indicated in cyan.

Simscape Multibody is used as a numerical reference model rather than as an independent experimental validation source. The Simscape and PACDM implementations share the mechanism geometry, joint definitions, fixed frame transformations, and prescribed active-coordinate inputs. In the PACDM implementation, however, the kinematic paths, closure residuals, PACDM Jacobians, and passive-coordinate solutions are constructed and evaluated independently through the proposed formulation. In particular, the passive-coordinate values obtained from Simscape are not supplied to the PACDM solver as initial estimates or correction data, but are retained only for post-solution comparison. The resulting comparison is therefore interpreted as numerical cross-validation between two kinematic solution procedures based on the same underlying mechanism description.

\subsection{Problem Formulation}
\begin{figure}[pos=htbp]
    \centering
    \includegraphics[width=\linewidth,trim={0cm 2.5cm 0cm 2.5cm},clip]{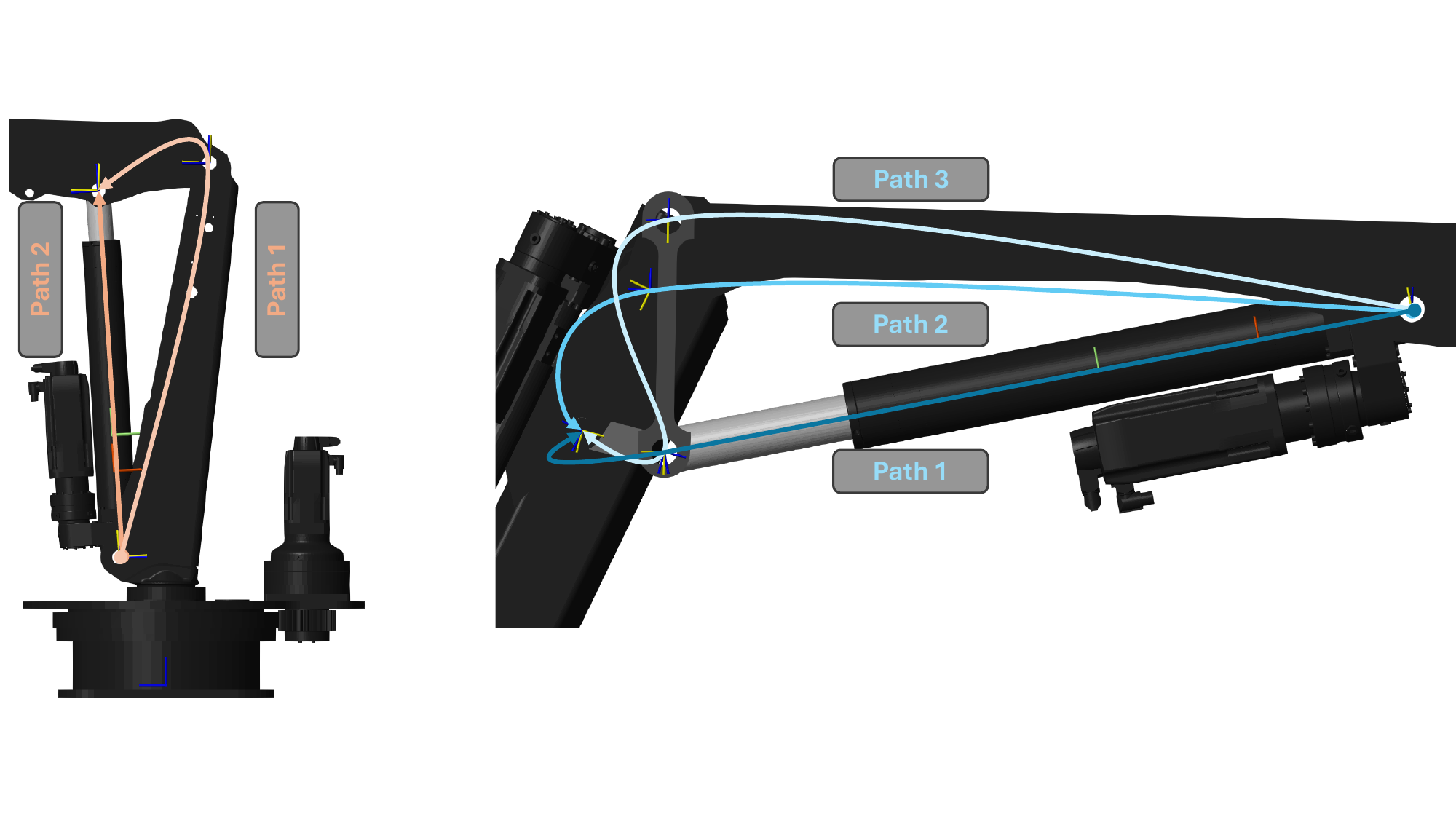}
    \caption{Closed-chain modules of the manipulator and their associated kinematic paths. \textit{left}: Module 1. \textit{right}: Module 2.}
    \label{fig:closed-chain-modules}
\end{figure}

\begin{figure}[pos=htbp]
    \centering
    \includegraphics[
        width=\linewidth,
        trim={0cm 5.75cm 0cm 4.75cm},
        clip
    ]{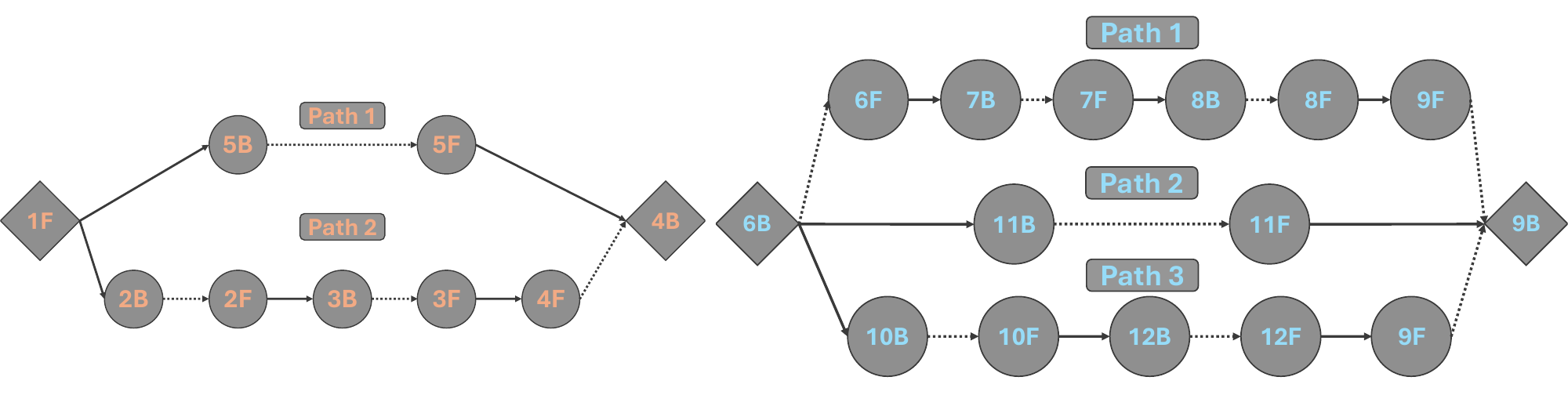}
    \caption{Local-transformation graphs associated with the kinematic paths of the manipulator's closed-chain modules. Rigid transforms within one link are represented by solid arrow lines, whereas the transforms from the joints' motion primitives are indicated by dashed arrow lines. The letters "B" and "F" denote the base and follower frames of joints, respectively. }
    \label{fig:path-transformation-graphs}
\end{figure}

The two closed-chain modules of this manipulator are shown in Fig. \ref{fig:closed-chain-modules}, each with its corresponding paths. The coordinate vectors of the first and second modules are, respectively,
\begin{equation}
 \widehat{\boldsymbol{q}}_1=\left[q_3 \; q_2 \; q_4 \; q_5\right]^T \qquad \text{and} \qquad \widehat{\boldsymbol{q}}_2=\left[q_7\;q_6 \;q_8 \;q_9 \;q_{10} \;q_{11} \;q_{12}\right]^T,
\end{equation}
with the active coordinates preceding the passive coordinates according to the active-passive partitions shown in Fig.~\ref{fig:simscape-robot-model}. The paths of the first module start from the follower frame of joint 1 and terminate at the base frame of joint 4. The paths of the second module start from the base frame of joint 6 and terminate at the base frame of joint 9. The graphs of the local transformations along each path are shown in Fig. \ref{fig:path-transformation-graphs}.

Accordingly, under the adopted multi-path formulation, the element-wise defected closure mismatch in \eqref{eq:defected-mismatch} specializes to
\begin{equation}
    \boldsymbol{\Lambda}_1\left(\widehat{\boldsymbol{q}}_1,\lambda\right)=\left[\boldsymbol{T}_1^{\langle2\rangle}\left(\widehat{\boldsymbol{q}}_1\right)\boldsymbol{D}_1\left(\lambda\right)\right]^{-1}\boldsymbol{T}_1^{\langle1\rangle}\left(\widehat{\boldsymbol{q}}_1\right),
\end{equation}
for the first module, and
\begin{equation}
        \begin{aligned}
        \boldsymbol{\Lambda}_1\left(\widehat{\boldsymbol{q}}_2,\lambda\right)&=\left[\boldsymbol{T}_1^{\langle2\rangle}\left(\widehat{\boldsymbol{q}}_2\right)\boldsymbol{D}_1\left(\lambda\right)\right]^{-1}\boldsymbol{T}_1^{\langle1\rangle}\left(\widehat{\boldsymbol{q}}_2\right),\\\boldsymbol{\Lambda}_2\left(\widehat{\boldsymbol{q}}_2,\lambda\right)&=\left[\boldsymbol{T}_2^{\langle3\rangle}\left(\widehat{\boldsymbol{q}}_2\right)\boldsymbol{D}_2\left(\lambda\right)\right]^{-1}\boldsymbol{T}_2^{\langle1\rangle}\left(\widehat{\boldsymbol{q}}_2\right),
    \end{aligned}
\end{equation}
for the second module. The tests in the following subsections are conducted using MATLAB\textregistered R2025b on a PC equipped with an Intel\textregistered Core\textcopyright i9-14900K CPU and a 64-bit Windows\texttrademark 11 operating system. Using the initial guesses, the rank of the passive block of the defected PACDM Jacobian for each module is determined using singular value decomposition (SVD), followed by column-pivoted QR (CPQR) factorization of its transpose to select the independent rows. The numerical parameter values used in this section are reported in Table \ref{tab:numerical-parameters}.

\begin{table}[pos=htbp]
    \centering
    \caption{Numerical parameters used in the proposed procedure.}
    \label{tab:numerical-parameters}
    \begin{tabular}{lll}
        \toprule
        \textbf{Method / procedure} & \textbf{Parameter} & \textbf{Value} \\
        \midrule

        Rank analysis
        & Relative SVD rank tolerance
        & $10^{-10}$ \\

        \addlinespace

        CPQR row selection
        & Selected passive-block reciprocal-condition threshold
        & $10^{-10}$ \\

        \addlinespace

        \multirow{3}{*}{Defect homotopy}
        & Initial continuation increment
        & $0.1$ \\
        & Minimum continuation increment
        & $10^{-5}$ \\
        & Failed-step reduction factor
        & $0.5$ \\

        \addlinespace

        \multirow{6}{*}{Homotopy corrector}
        & Residual tolerance
        & $10^{-9}$ \\
        & Step-stagnation tolerance
        & $10^{-12}$ \\
        & Maximum iterations
        & $100$ \\
        & Initial damping parameter
        & $10^{-6}$ \\
        & Damping-parameter bounds
        & $[10^{-12},\,10^{12}]$ \\
        & Damping update factor
        & $10$ \\

        \addlinespace

        \multirow{2}{*}{Closure acceptance}
        & Selected physical-residual tolerance
        & $10^{-9}$ \\
        & Complete raw defect/closure residual tolerance
        & $10^{-8}$ \\

        \addlinespace

        \multirow{6}{*}{Trajectory corrector}
        & Physical-corrector residual tolerance
        & $10^{-9}$ \\
        & Maximum direct-corrector iterations
        & $8$ \\
        & Maximum line-search iterations
        & $4$ \\
        & Passive-block reciprocal-condition threshold
        & $10^{-11}$ \\
        & Fallback Gauss--Newton damping
        & $10^{-8}$ \\
        & Negligible trial-step tolerance
        & $10^{-13}$ \\

        \bottomrule
    \end{tabular}
\end{table}

The relative SVD rank tolerance and the CPQR selected passive-block reciprocal-condition threshold are independent numerical criteria, although both are assigned the value $10^{-10}$ in the reported simulations. The former determines the numerical rank from relative singular values, whereas the latter is applied separately to reject a selected passive block that is regarded as numerically ill-conditioned. The trajectory corrector employs the separate reciprocal-condition threshold listed in Table \ref{tab:numerical-parameters}. These tolerance values are implementation choices, and no invariance with respect to numerical scaling or tolerance selection is claimed.

\subsection{Fixed-Configuration Test}
\begin{figure}[pos=htbp]
    \centering
    \includegraphics[
        width=\linewidth,
        trim={0.1cm 0.1cm 0.1cm 0.1cm},
        clip
    ]{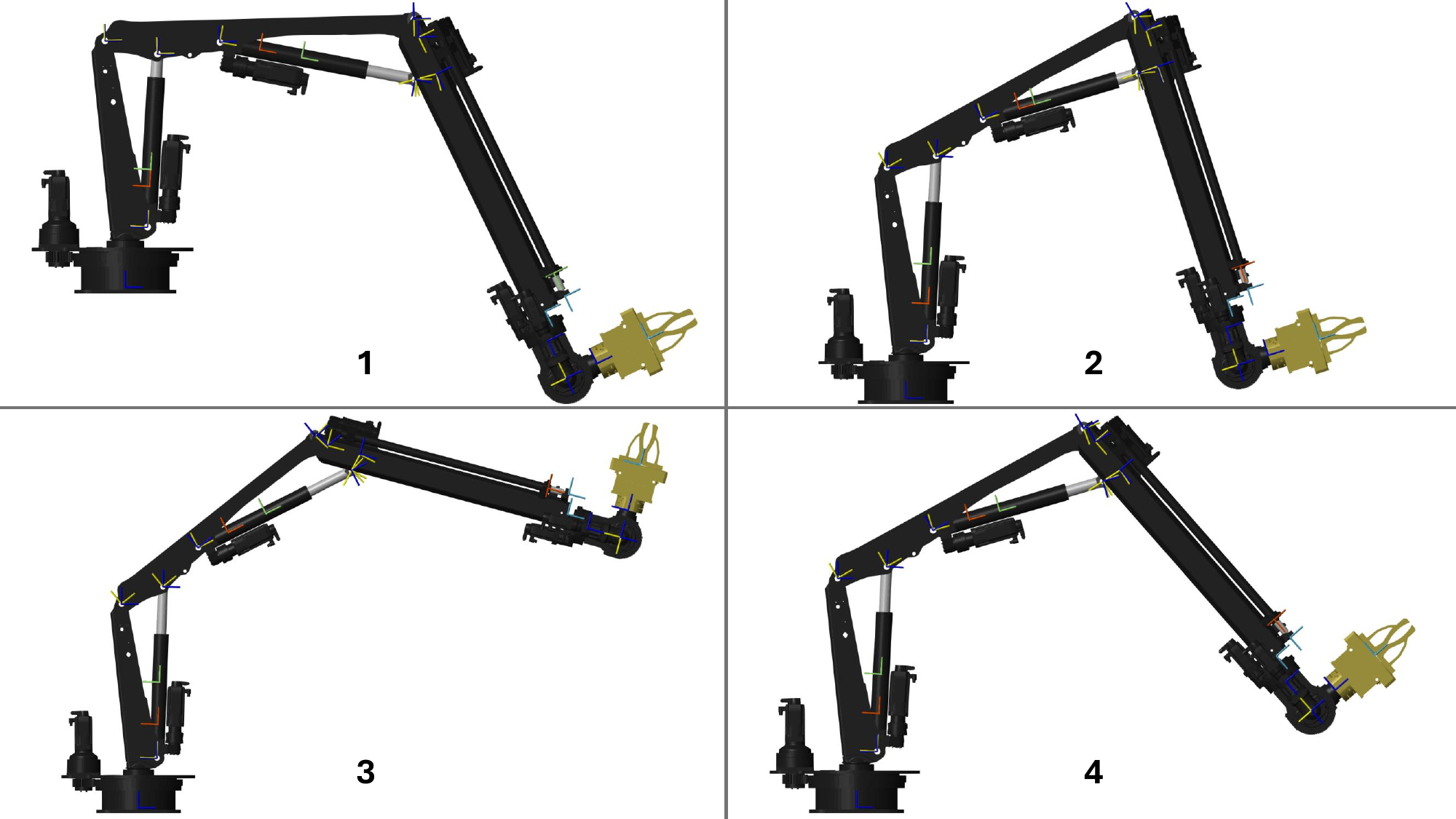}
    \caption{Four fixed configurations of the manipulator corresponding to the prescribed active-coordinate values used in the configuration-level validation test.}
    \label{fig:fixed-test-configurations}
\end{figure}

In the first test, we prescribe values for the active coordinates at different fixed configurations and acquire the passive coordinates via defect homotopy. The numerical results are then compared with reference values from Simscape Multibody. As mentioned in Section \ref{sec:defect-homotopy}, the proposed framework can be initialized with a rough guess, which can be obtained, for instance, by visual inspection of the current configuration. The four fixed configurations for this test are shown in Fig. \ref{fig:fixed-test-configurations}. Accordingly, the selected initial guesses based on visual inspection of the configurations are reported in Table \ref{tab:fixed-configuration-initialization}, along with the active-coordinate values.

\begin{table}[pos=htbp]
    \centering
    \caption{Active-coordinate values and passive-coordinate initial guesses used in the fixed-configuration tests. Angular coordinates are reported in degrees, while the prismatic coordinates $q_3$ and $q_7$ are reported in meters.}
    \label{tab:fixed-configuration-initialization}
    \renewcommand{\arraystretch}{1.35}
    \setlength{\tabcolsep}{5pt}
    \begin{tabular}{c c c c}
        \toprule
        \textbf{Test}
        &
        \begin{tabular}[c]{@{}c@{}}
            $\boldsymbol{q}_a$\\
            $(q_1,q_3,q_7,q_{13},q_{14},q_{15},q_{16})$
        \end{tabular}
        &
        \begin{tabular}[c]{@{}c@{}}
            $\widehat{\boldsymbol q}_{1,p}^{\,0}$ (deg)\\
            $(q_2,q_4,q_5)$
        \end{tabular}
        &
        \begin{tabular}[c]{@{}c@{}}
            $\widehat{\boldsymbol q}_{2,p}^{\,0}$ (deg)\\
            $(q_6,q_8,q_9,q_{10},q_{11},q_{12})$
        \end{tabular}
        \\
        \midrule

        1
        &
        $\begin{bmatrix}
            0 & 0.10 & 0.25 & 0 & 0 & 180 & 0
        \end{bmatrix}^{\mathsf T}$
        &
        $\begin{bmatrix}
            0 & 0 & 0
        \end{bmatrix}^{\mathsf T}$
        &
        $\begin{bmatrix}
            0 & -90 & -90 & -90 & -45 & 90
        \end{bmatrix}^{\mathsf T}$
        \\[1ex]

        2
        &
        $\begin{bmatrix}
            0 & 0.25 & 0.10 & 0 & 0 & 180 & 0
        \end{bmatrix}^{\mathsf T}$
        &
        $\begin{bmatrix}
            0 & 30 & 30
        \end{bmatrix}^{\mathsf T}$
        &
        $\begin{bmatrix}
            -10 & -90 & -90 & -100 & -90 & 90
        \end{bmatrix}^{\mathsf T}$
        \\[1ex]

        3
        &
        $\begin{bmatrix}
            0 & 0.30 & 0.30 & 0 & 0 & 180 & 0
        \end{bmatrix}^{\mathsf T}$
        &
        $\begin{bmatrix}
            0 & 45 & 45
        \end{bmatrix}^{\mathsf T}$
        &
        $\begin{bmatrix}
            -30 & -45 & -90 & -90 & -60 & 90
        \end{bmatrix}^{\mathsf T}$
        \\[1ex]

        4
        &
        $\begin{bmatrix}
            0 & 0.25 & 0.20 & 0 & 0 & 180 & 0
        \end{bmatrix}^{\mathsf T}$
        &
        $\begin{bmatrix}
            0 & 30 & 30
        \end{bmatrix}^{\mathsf T}$
        &
        $\begin{bmatrix}
            -10 & -90 & -90 & -90 & -90 & 90
        \end{bmatrix}^{\mathsf T}$
        \\

        \bottomrule
    \end{tabular}
\end{table}

 The resulting acquisition and physical passive-block ranks and selected residual-row indices for each module are listed in Table \ref{tab:passive-ranks-selected-rows}. In all four fixed-configuration tests, the passive-block rank equals the number of passive coordinates for both modules. The retained row-index sets are $\{3,4,5\}$ for Module 1 and $\{3,4,5,9,10,11\}$ for Module 2 at both the acquisition and physical configurations, although the CPQR pivot ordering varies in some Module 2 cases. The passive coordinates obtained through defect homotopy, along with the reference values from Simscape Multibody and the corresponding errors, are shown in Table \ref{tab:fixed-passive-comparison}.

\begin{table}[pos=htbp]
    \centering
    \caption{Passive-block ranks and selected residual-row indices for the fixed-configuration tests. Acquisition quantities are evaluated from the defected passive Jacobian at the artificial starting configuration $(\widehat{\boldsymbol{q}}_m^0,0)$, whereas physical quantities are evaluated after recovery of the physical closure configuration at $\lambda=1$. The same passive-block ranks and retained row-index sets were obtained in all four tests; the ordering of the CPQR pivots may differ.}
    \label{tab:passive-ranks-selected-rows}
    \renewcommand{\arraystretch}{1.2}
    \setlength{\tabcolsep}{4pt}
    \begin{tabular}{c c c c c c}
        \toprule
        \textbf{Module}
        & \textbf{$n_{m,p}$}
        & \begin{tabular}[c]{@{}c@{}}
              \textbf{Acquisition}\\
              \textbf{passive rank}
          \end{tabular}
        & \begin{tabular}[c]{@{}c@{}}
              \textbf{Acquisition}\\
              \textbf{rows $\boldsymbol{S}_m^{0}$}
          \end{tabular}
        & \begin{tabular}[c]{@{}c@{}}
              \textbf{Physical}\\
              \textbf{passive rank}
          \end{tabular}
        & \begin{tabular}[c]{@{}c@{}}
              \textbf{Physical}\\
              \textbf{rows $\boldsymbol{S}_m$}
          \end{tabular}
        \\
        \midrule

        1
        & 3
        & 3
        & $\{3,4,5\}$
        & 3
        & $\{3,4,5\}$ \\

        2
        & 6
        & 6
        & $\{3,4,5,9,10,11\}$
        & 6
        & $\{3,4,5,9,10,11\}$ \\

        \bottomrule
    \end{tabular}
\end{table}

\begin{table}[pos=htbp]
    \centering
    \caption{Passive coordinates obtained through defect homotopy and the
    corresponding Simscape Multibody values for the fixed-configuration tests. The reported values are in radians.}
    \label{tab:fixed-passive-comparison}
    \renewcommand{\arraystretch}{1.1}
    \setlength{\tabcolsep}{4pt}
    \begin{tabular}{cccrrr}
        \toprule
        \textbf{Module}
        & \textbf{Test}
        & \textbf{Coordinate}
        & \textbf{Simscape}
        & \textbf{Calculated}
        & \textbf{Abs. error} \\
        \midrule

        \multirow{12}{*}{1}

        & \multirow{3}{*}{1}
        & $q_2$ & $-0.057312$ & $-0.057312$ & $2.776\times10^{-17}$ \\
        &       & $q_4$ & $ 0.09301 $ & $ 0.09301 $ & $5.246\times10^{-15}$ \\
        &       & $q_5$ & $ 0.035698$ & $ 0.035698$ & $4.288\times10^{-15}$ \\

        \cmidrule(lr){2-6}

        & \multirow{3}{*}{2}
        & $q_2$ & $-0.051033$ & $-0.051033$ & $6.938\times10^{-11}$ \\
        &       & $q_4$ & $ 0.56342 $ & $ 0.56342 $ & $1.150\times10^{-10}$ \\
        &       & $q_5$ & $ 0.51239 $ & $ 0.51239 $ & $4.778\times10^{-11}$ \\

        \cmidrule(lr){2-6}

        & \multirow{3}{*}{3}
        & $q_2$ & $-0.035406$ & $-0.035406$ & $1.312\times10^{-15}$ \\
        &       & $q_4$ & $ 0.71581 $ & $ 0.71581 $ & $7.550\times10^{-15}$ \\
        &       & $q_5$ & $ 0.6804  $ & $ 0.6804  $ & $7.328\times10^{-15}$ \\

        \cmidrule(lr){2-6}

        & \multirow{3}{*}{4}
        & $q_2$ & $-0.051033$ & $-0.051033$ & $6.938\times10^{-11}$ \\
        &       & $q_4$ & $ 0.56342 $ & $ 0.56342 $ & $1.150\times10^{-10}$ \\
        &       & $q_5$ & $ 0.51239 $ & $ 0.51239 $ & $4.778\times10^{-11}$ \\

        \midrule

        \multirow{24}{*}{2}

        & \multirow{6}{*}{1}
        & $q_6$    & $-0.22439$ & $-0.22439$ & $2.497\times10^{-12}$ \\
        &          & $q_8$    & $-1.13$    & $-1.13$    & $5.538\times10^{-11}$ \\
        &          & $q_9$    & $-1.7693$  & $-1.7693$  & $1.607\times10^{-11}$ \\
        &          & $q_{10}$ & $-1.5888$  & $-1.5888$  & $1.431\times10^{-11}$ \\
        &          & $q_{11}$ & $-1.1559$  & $-1.1559$  & $3.694\times10^{-11}$ \\
        &          & $q_{12}$ & $ 1.3364$  & $ 1.3364$  & $6.690\times10^{-11}$ \\

        \cmidrule(lr){2-6}

        & \multirow{6}{*}{2}
        & $q_6$    & $-0.22343$ & $-0.22343$ & $2.165\times10^{-15}$ \\
        &          & $q_8$    & $-1.5167$  & $-1.5167$  & $7.283\times10^{-14}$ \\
        &          & $q_9$    & $-1.5382$  & $-1.5382$  & $1.910\times10^{-14}$ \\
        &          & $q_{10}$ & $-2.0146$  & $-2.0146$  & $4.441\times10^{-15}$ \\
        &          & $q_{11}$ & $-1.7727$  & $-1.7727$  & $8.460\times10^{-14}$ \\
        &          & $q_{12}$ & $ 1.2963$  & $ 1.2963$  & $7.661\times10^{-14}$ \\

        \cmidrule(lr){2-6}

        & \multirow{6}{*}{3}
        & $q_6$    & $-0.21263$ & $-0.21263$ & $3.853\times10^{-13}$ \\
        &          & $q_8$    & $-0.95748$ & $-0.95748$ & $6.742\times10^{-12}$ \\
        &          & $q_9$    & $-1.7997$  & $-1.7997$  & $6.569\times10^{-12}$ \\
        &          & $q_{10}$ & $-1.4424$  & $-1.4424$  & $9.108\times10^{-13}$ \\
        &          & $q_{11}$ & $-0.94118$ & $-0.94118$ & $1.931\times10^{-13}$ \\   
        &          & $q_{12}$ & $ 1.2985$  & $ 1.2985$  & $7.231\times10^{-12}$ \\

        \cmidrule(lr){2-6}

        & \multirow{6}{*}{4}
        & $q_6$    & $-0.23047$ & $-0.23047$ & $1.141\times10^{-13}$ \\
        &          & $q_8$    & $-1.2775$  & $-1.2775$  & $1.360\times10^{-12}$ \\
        &          & $q_9$    & $-1.7146$  & $-1.7146$  & $1.153\times10^{-12}$ \\
        &          & $q_{10}$ & $-1.7311$  & $-1.7311$  & $1.799\times10^{-14}$ \\
        &          & $q_{11}$ & $-1.3642$  & $-1.3642$  & $3.080\times10^{-13}$ \\
        &          & $q_{12}$ & $ 1.3477$  & $ 1.3477$  & $1.498\times10^{-12}$ \\

        \bottomrule
    \end{tabular}
\end{table}

The results in Table \ref{tab:fixed-passive-comparison} show close agreement between the passive coordinates obtained through defect homotopy and the corresponding Simscape Multibody values for all four fixed configurations. The largest absolute error among the reported coordinates is approximately $1.150\times10^{-10}$, while several cases exhibit errors close to machine precision. Comparable accuracy is obtained for both closed-chain modules despite their different path structures, with the three-path second module maintaining absolute errors below $7\times10^{-11}$ across the tested configurations. In all four tests, the defect-homotopy procedure therefore recovers closure-consistent passive coordinates associated with the Simscape reference configurations from the prescribed approximate initial estimates.

\subsection{Continuous-Trajectory Test}
After demonstrating the ability of the proposed method to acquire closure-consistent configurations from rough initial guesses, we use the predictor \eqref{eq:continuation-predictor} and the corrector \eqref{eq:continuation-corrector} in this test to obtain the configurations resulting from a prescribed continuous trajectory for the active coordinates. Defect homotopy may alternatively be applied at every trajectory step, but doing so entails substantially greater computational cost than the local predictor--corrector continuation. This computational difference is also investigated in this subsection.

\begin{figure}[pos=htbp]
    \centering
    \includegraphics[
        width=\linewidth,
        trim={0cm 0cm 0cm 0cm},
        clip
    ]{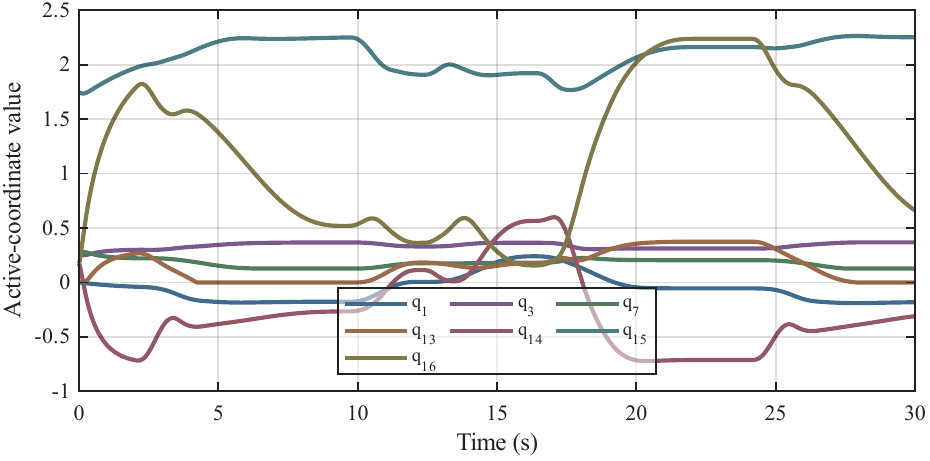}
    \caption{Prescribed active-coordinate trajectory used for the continuous-motion validation test. Values are in radians except for $q_3$, $q_7$ and $q_{13}$ which are in meters. The sampling time in this test is 1ms.}
    \label{fig:active-coordinate-trajectory}
\end{figure}

\begin{figure}[pos=htbp]
    \centering
    \includegraphics[
        width=\linewidth,
        trim={0cm 0cm 0cm 0cm},
        clip
    ]{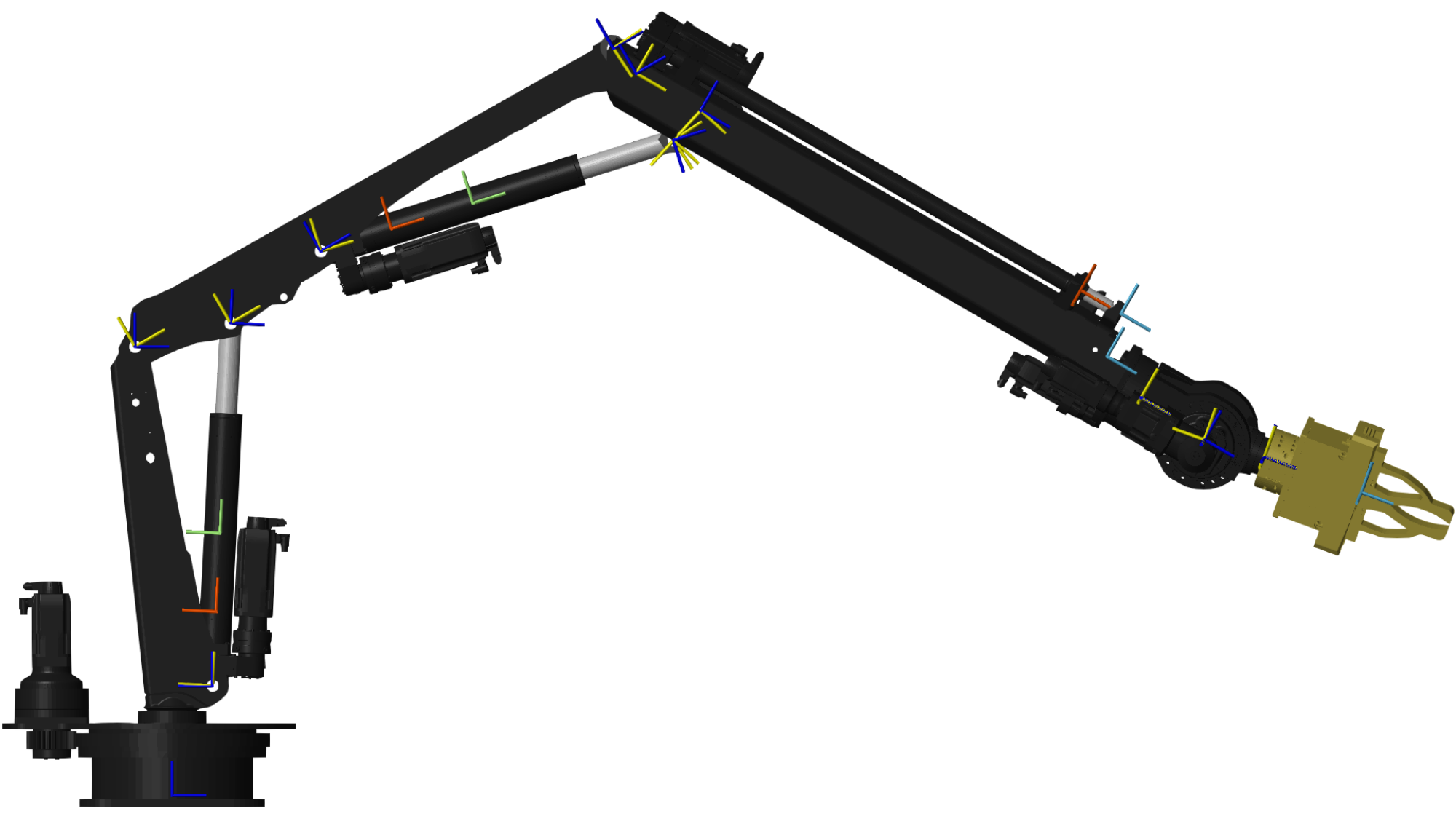}
    \caption{Initial configuration of the manipulator for the continuous-trajectory validation test.}
    \label{fig:trajectory-initial-configuration}
\end{figure}

\begin{figure}[pos=htbp]
    \centering
    \captionsetup[subfigure]{labelformat=empty}

    \begin{subfigure}[t]{0.85\linewidth}
        \centering
        \includegraphics[
            width=\linewidth,
            trim={0cm 0cm 0cm 0cm},
            clip
        ]{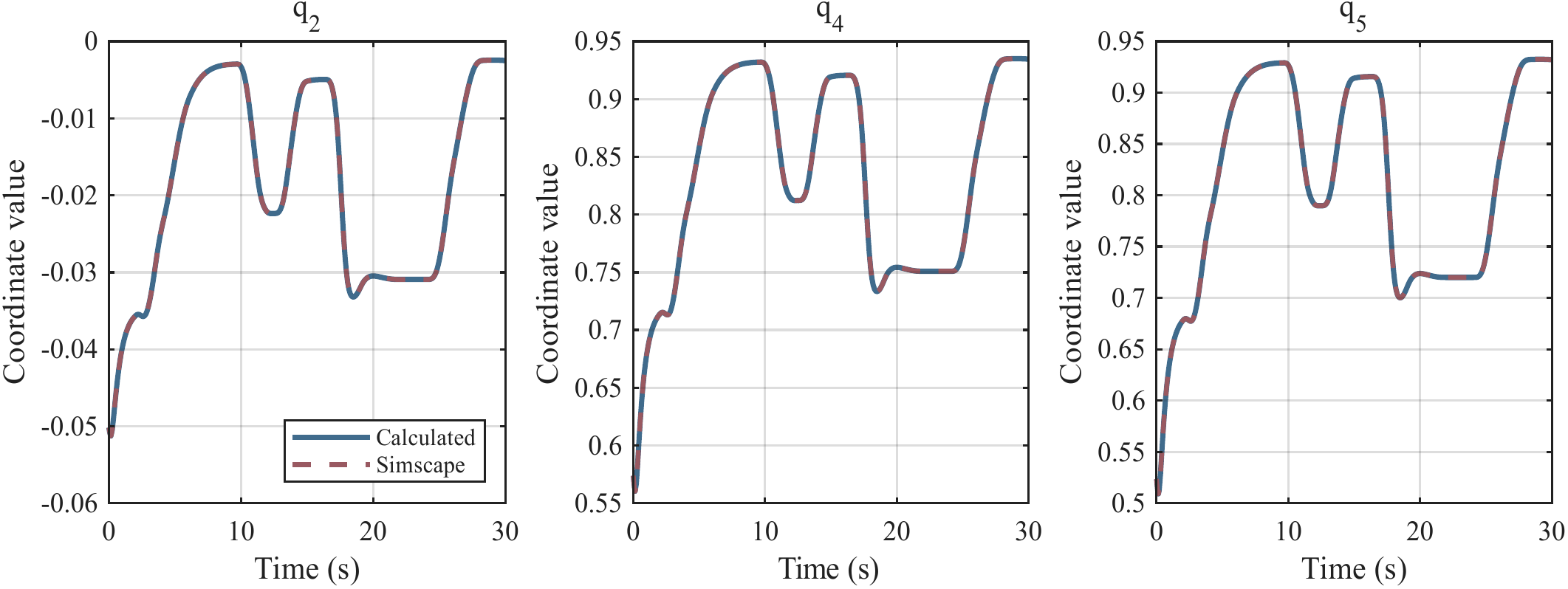}
        \caption*{Module 1}
    \end{subfigure}

    \vspace{0.8em}

    \begin{subfigure}[t]{0.85\linewidth}
        \centering
        \includegraphics[
            width=\linewidth,
            trim={0cm 0cm 0cm 0cm},
            clip
        ]{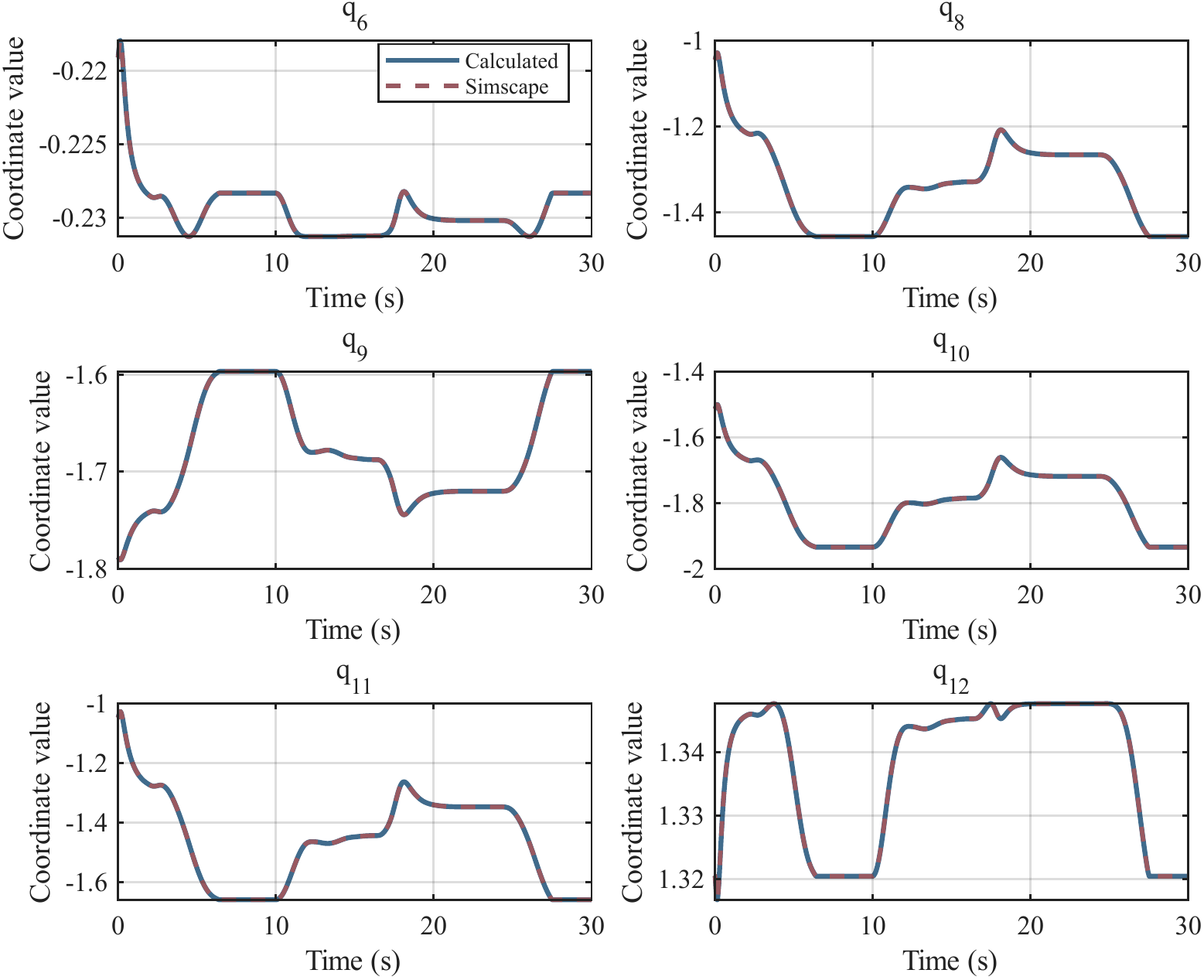}
        \caption*{Module 2}
    \end{subfigure}

    \caption{Comparison of the passive-coordinate trajectories obtained along the prescribed motion with the corresponding reference values from Simscape Multibody for the two closed-chain modules. The values are in radians.}
    \label{fig:passive-trajectory-comparison}
\end{figure}

For the prescribed active-coordinate trajectory displayed in Fig. \ref{fig:active-coordinate-trajectory}, the initial configuration is shown in Fig. \ref{fig:trajectory-initial-configuration}. The rough initial guess obtained by visual inspection of this configuration is reported in Table \ref{tab:trajectory-initial-guess}. The passive coordinates obtained along the prescribed trajectory are compared in Fig. \ref{fig:passive-trajectory-comparison} with the corresponding reference values from Simscape Multibody. The root-mean-square error (RMSE), maximum absolute error, and 95th-percentile absolute error are reported for each module in Table \ref{tab:trajectory-error-metrics}. The reported RMSE is calculated by pooling the scalar passive-coordinate errors over all passive coordinates and all trajectory samples within each module. The passive-coordinate trajectories obtained using the predictor--corrector continuation remain in close agreement with the Simscape Multibody reference throughout the prescribed motion.

As summarized in Table~\ref{tab:trajectory-error-metrics}, the RMSE is $8.4702\times10^{-10}$ for Module 1 and $7.3249\times10^{-10}$ for Module 2, while the maximum absolute errors remain below $3.1\times10^{-9}$. The 95th-percentile absolute errors are below $2.0\times10^{-9}$ for both modules, indicating that the small discrepancies are maintained over the trajectory rather than being limited to isolated configurations. The similar error levels obtained for the two modules further show that the local first-order prediction followed by direct closure correction can maintain the passive-coordinate solution along the prescribed continuous motion for both the two-path and three-path module structures considered here.

\begin{table}[pos=htbp]
    \centering
    \caption{Passive-coordinate initial guesses used for the
    continuous-trajectory test.}
    \label{tab:trajectory-initial-guess}
    \renewcommand{\arraystretch}{1.1}
    \begin{tabular}{ccc}
        \toprule
        \textbf{Module}
        & \textbf{Passive coordinate}
        & \textbf{Initial guess (deg)} \\
        \midrule

        \multirow{3}{*}{1}
        & $q_2$ & $0$ \\
        & $q_4$ & $30$ \\
        & $q_5$ & $30$ \\

        \midrule

        \multirow{6}{*}{2}
        & $q_6$    & $-10$ \\
        & $q_8$    & $-60$ \\
        & $q_9$    & $-90$ \\
        & $q_{10}$ & $-90$ \\
        & $q_{11}$ & $-60$ \\
        & $q_{12}$ & $60$ \\

        \bottomrule
    \end{tabular}
\end{table}

\begin{table}[pos=htbp]
    \centering
    \caption{Passive-coordinate error metrics for the continuous-trajectory test. Reported values are in radians.}
    \label{tab:trajectory-error-metrics}
    \renewcommand{\arraystretch}{1.1}
    \begin{tabular}{cccc}
        \toprule
        \textbf{Module}
        & \textbf{RMSE}
        & \textbf{Maximum abs. error}
        & \textbf{95th-percentile abs. error} \\
        \midrule
        1
        & $8.470197\times10^{-10}$
        & $2.733605\times10^{-9}$
        & $1.826391\times10^{-9}$ \\
        2
        & $7.324859\times10^{-10}$
        & $3.065632\times10^{-9}$
        & $1.922650\times10^{-9}$ \\
        \bottomrule
    \end{tabular}
\end{table}

A corresponding comparison of the passive-coordinate errors is listed in Table \ref{tab:trajectory-method-comparison}, along with the average runtime over five repeated runs, for predictor--corrector continuation following the initial defect-homotopy acquisition and for defect homotopy applied at every trajectory step. The comparison in Table~\ref{tab:trajectory-method-comparison} shows that both strategies retain comparable numerical accuracy. Defect homotopy at every trajectory step yields slightly smaller RMSE and 95th-percentile absolute errors for both modules, whereas the predictor--corrector approach yields smaller maximum absolute errors for both modules. The RMSE values remain of the $10^{-10}$ order for both strategies, while the maximum absolute errors remain below $5\times10^{-9}$.

The more significant difference is computational: the average runtime decreases from $980.542\,\mathrm{s}$ for defect homotopy applied at every trajectory step to $21.411\,\mathrm{s}$ for predictor--corrector continuation following the initial defect-homotopy acquisition. This corresponds to an approximately $45.80$-fold reduction in runtime while retaining comparable numerical accuracy. For the tested trajectory, the results therefore support using defect homotopy for the initial passive-coordinate branch acquisition and the local predictor--corrector procedure for subsequent continuation along nearby configurations.

The numerical study exercises both the elementary two-path construction and its extension to a multi-path module within the considered multibody system. The reported results are limited to the tested fixed configurations and prescribed trajectory and are intended to verify the proposed formulation and its numerical implementation for these cases. They do not constitute a statistical assessment of robustness over arbitrary closed-chain mechanisms, initial conditions, or operating configurations.

\begin{table}[pos=htbp]
    \centering
    \caption{Comparison of passive-coordinate errors and computational
    runtime for the two trajectory-continuation strategies. Runtime values
    are averaged over five repeated runs. Reported error metrics are in radians.}
    \label{tab:trajectory-method-comparison}
    \renewcommand{\arraystretch}{1.2}
    \setlength{\tabcolsep}{4pt}
    \begin{tabular}{c c c c c c}
        \toprule
        \textbf{Method}
        & \textbf{Module}
        & \textbf{RMSE}
        & \begin{tabular}[c]{@{}c@{}}
              \textbf{Maximum}\\
              \textbf{abs. error}
          \end{tabular}
        & \begin{tabular}[c]{@{}c@{}}
              \textbf{95th-percentile}\\
              \textbf{abs. error}
          \end{tabular}
        & \begin{tabular}[c]{@{}c@{}}
              \textbf{Average}\\
              \textbf{runtime (s)}
          \end{tabular}
        \\
        \midrule

        \multirow{2}{*}{\begin{tabular}[c]{@{}c@{}}
            Predictor--corrector\\
            continuation
        \end{tabular}}
        & 1
        & $8.4702\times10^{-10}$
        & $2.7336\times10^{-9}$
        & $1.8264\times10^{-9}$
        & \multirow{2}{*}{$21.411$}
        \\

        & 2
        & $7.3249\times10^{-10}$
        & $3.0656\times10^{-9}$
        & $1.9227\times10^{-9}$
        &
        \\

        \midrule

        \multirow{2}{*}{\begin{tabular}[c]{@{}c@{}}
            Defect homotopy\\
            at every step
        \end{tabular}}
        & 1
        & $5.8145\times10^{-10}$
        & $4.9028\times10^{-9}$
        & $4.8709\times10^{-10}$
        & \multirow{2}{*}{$980.542$}
        \\

        & 2
        & $6.6785\times10^{-10}$
        & $4.5887\times10^{-9}$
        & $1.8485\times10^{-9}$
        &
        \\

        \bottomrule
    \end{tabular}
\end{table}

\section{Conclusion}
\label{sec:conclusion}

This paper presented a path-assembled closure differential mapping framework for the modular kinematic treatment of mechanisms containing closed kinematic chains. The framework constructs closure relations directly from ordered transformation paths with common endpoints and represents the resulting path mismatch through local logarithmic coordinates in $\mathrm{SE}(3)$. Differentiation of this residual yields the Path-Assembled Closure Differential Mapping (PACDM), whose element-level contributions can be assembled in a common module-coordinate space. For modules containing several paths, a graph-theoretically minimal set of two-path closure elements is sufficient to impose path equality, while rank-revealing analysis of the assembled differential identifies the locally independent scalar closure conditions. Consequently, the formulation avoids the mechanism-specific derivation of minimal scalar closure equations and does not assume that every component of the raw closure residual is independent.

A defect-homotopy procedure was introduced to recover closure-consistent passive coordinates from prescribed active coordinates and an approximate passive-coordinate estimate. The initial path mismatch is incorporated as a defect so that the estimated configuration becomes an exact solution of an artificial closure problem. Continuation then removes this defect and traces the connected solution branch to the physical closure equations. Residual coordinates used during branch acquisition are selected from the defected passive Jacobian at the artificial starting configuration. After the physical configuration is recovered, the complete physical closure residual and the relevant Jacobian ranks are verified independently, and the residual coordinates used for kinematic reduction are selected anew from the physical passive-coordinate block. This separation prevents the artificial starting problem from determining the independent physical closure representation.

At a regular physical configuration, partitioning the selected PACDM Jacobian into active- and passive-coordinate blocks provides the local active-to-passive differential mapping through the implicit function theorem. The resulting intermediate Jacobian expresses variations of all module coordinates in terms of the active-coordinate variations and therefore supplies a compact kinematic interface for subsequent modular modeling or control formulations. Once the desired physical branch has been acquired, this local mapping can also be used as a first-order predictor during continuous motion, followed by direct correction of the physical closure equations. Defect homotopy can then be retained as a branch-recovery procedure when direct correction is unsuccessful.

The numerical study considered a 7-DoF heavy-duty robotic manipulator with two closed-chain modules comprising two and three kinematic paths. In all four fixed-configuration tests, the passive-coordinate Jacobian blocks attained the required ranks, and the calculated passive coordinates agreed closely with the Simscape Multibody reference values. The largest reported absolute error was approximately $1.15 \times 10^{-10}$. Along the prescribed continuous trajectory, the predictor--corrector procedure produced root-mean-square errors of $8.4702 \times 10^{-10}$ and $7.3249 \times 10^{-10}$ for the two modules, respectively, while the maximum absolute errors remained below $3.1 \times 10^{-9}$. Its average runtime was $21.411~\mathrm{s}$, compared with $980.542~\mathrm{s}$ when defect homotopy was applied at every trajectory step. This corresponds to an approximately $45.8$-fold reduction in runtime while retaining comparable numerical accuracy.

The reported results verify the proposed formulation and its numerical implementation for the tested configurations and trajectory, but they do not establish global convergence or statistical robustness across arbitrary closed-chain mechanisms. The method remains local to the selected logarithm branch and requires the intended passive-coordinate branch to remain within the prescribed coordinate bounds, retain a nonsingular selected passive Jacobian, and avoid branch boundaries or other losses of regularity. Future work will consider automated module and path extraction from mechanism graphs, adaptive treatment of logarithm-chart boundaries and changes in constraint rank, broader evaluation across different closed-chain topologies, and integration of the resulting reduced mappings into recursive dynamics and modular control frameworks.

\appendix

\section{Construction of Local Homogeneous Transformations}
\label{app:local-transformations}

The path transformations introduced in Subsection \ref{subsec:two-path-closure} are assembled from local homogeneous transformations between consecutive frames along each selected kinematic path. Each local transformation represents either a fixed geometric relation between two frames or the relative motion introduced by a joint. Homogeneous transformations, their composition, and their exponential-coordinate representation are standard constructions in robot kinematics~\cite{lynch2017}.

Let $\mathcal{F}_i$ and $\mathcal{F}_j$ denote two consecutive frames, and let $\boldsymbol{\mathcal{T}}_{ij}\in\mathrm{SE}(3)$ map coordinates expressed in $\mathcal{F}_j$ to coordinates expressed in $\mathcal{F}_i$. The transformation is written as
\begin{equation}
    \boldsymbol{\mathcal{T}}_{ij}
    =
    \begin{bmatrix}
        \boldsymbol{R}_{ij} & \boldsymbol{p}_{ij} \\[2pt]
        \boldsymbol{0}_3^T & 1
    \end{bmatrix},
    \label{eq:appendix-homogeneous-transform}
\end{equation}
where $\boldsymbol{R}_{ij}\in\mathrm{SO}(3)$ describes the orientation of $\mathcal{F}_j$ relative to $\mathcal{F}_i$, and $\boldsymbol{p}_{ij}\in\mathbb{R}^3$ gives the position of the origin of $\mathcal{F}_j$ expressed in $\mathcal{F}_i$. With this convention, successive transformations are composed according to
\begin{equation}
    \boldsymbol{\mathcal{T}}_{ik}
    =
    \boldsymbol{\mathcal{T}}_{ij}\boldsymbol{\mathcal{T}}_{jk}.
    \label{eq:appendix-transform-composition}
\end{equation}
If a geometric relation is traversed in the direction opposite to that in which its transformation is defined, the corresponding inverse transformation is used.

A fixed local transformation is independent of the generalized coordinates and is therefore obtained directly from the constant relative rotation and translation between its adjacent frames. Denoting such a factor by $\boldsymbol{\mathcal{T}}_{F}$ gives
\begin{equation}
    \boldsymbol{\mathcal{T}}_{F}
    =
    \begin{bmatrix}
        \boldsymbol{R}_{F} & \boldsymbol{p}_{F} \\[2pt]
        \boldsymbol{0}_3^T & 1
    \end{bmatrix}.
    \label{eq:appendix-fixed-transform}
\end{equation}
Consequently, its coordinate derivative vanishes.

Joint-dependent local transformations are represented through the exponential map. Let $q_j$ denote the generalized coordinate associated with an elementary joint factor and let $\boldsymbol{\xi}_j\in\mathbb{R}^6$ denote its local motion vector under the adopted vee convention. The corresponding transformation is
\begin{equation}
    \boldsymbol{\mathcal{T}}_{J,j}(q_j)
    =
    \operatorname{Exp}
    \left(
        s_j\boldsymbol{\xi}_j q_j
    \right),
    \qquad
    s_j\in\{-1,+1\},
    \label{eq:appendix-joint-exp}
\end{equation}
where $s_j$ accounts for the adopted coordinate direction along the selected path. Here, $\operatorname{Exp}:\mathbb{R}^{6}\rightarrow\mathrm{SE}(3)$ denotes the Lie-group exponential associated with the matrix exponential of the corresponding element of $\mathfrak{se}(3)$.

The joint frames may be chosen such that the local $z$-axis coincides with the joint axis. Let
\begin{equation}
    \boldsymbol{e}_z
    =
    \begin{bmatrix}
        0 & 0 & 1
    \end{bmatrix}^{T}.
\end{equation}
Under this frame convention, the local motion vectors for revolute and prismatic joints are, respectively,
\begin{equation}
    \boldsymbol{\xi}_{r}
    =
    \operatorname{col}
    \left(
        \boldsymbol{e}_z,\boldsymbol{0}_3
    \right),
    \qquad
    \boldsymbol{\xi}_{p}
    =
    \operatorname{col}
    \left(
        \boldsymbol{0}_3,\boldsymbol{e}_z
    \right).
    \label{eq:appendix-motion-vectors}
\end{equation}
The associated elementary transformations can therefore be written explicitly as
\begin{equation}
    \boldsymbol{\mathcal{T}}_{r}(q_j)
    =
    \begin{bmatrix}
        \boldsymbol{R}_{z}(s_jq_j) & \boldsymbol{0}_3 \\[3pt]
        \boldsymbol{0}_3^T & 1
    \end{bmatrix},
    \label{eq:appendix-revolute-transform}
\end{equation}
and
\begin{equation}
    \boldsymbol{\mathcal{T}}_{p}(q_j)
    =
    \begin{bmatrix}
        \boldsymbol{I}_3 & s_jq_j\boldsymbol{e}_z \\[2pt]
        \boldsymbol{0}_3^T & 1
    \end{bmatrix},
    \label{eq:appendix-prismatic-transform}
\end{equation}
where
\begin{equation}
    \boldsymbol{R}_{z}(\theta)
    =
    \begin{bmatrix}
        \cos\theta & -\sin\theta & 0 \\
        \sin\theta &  \cos\theta & 0 \\
        0          &  0          & 1
    \end{bmatrix}.
    \label{eq:appendix-rz}
\end{equation}
The use of a local $z$-axis motion primitive does not restrict the orientation or location of the physical joint axis. Arbitrary joint geometry is represented by the surrounding fixed transformations, whose frames are positioned and oriented such that the corresponding elementary joint motion is expressed in the joint-aligned local frame.

This construction also gives the local right-trivialized derivatives used in Subsection 2.2. For a fixed transformation,
\begin{equation}
    \left[
        \frac{\partial \boldsymbol{\mathcal{T}}_{F}}
             {\partial q_j}
        \boldsymbol{\mathcal{T}}_{F}^{-1}
    \right]^\vee
    =
    \boldsymbol{0}_6,
    \label{eq:appendix-fixed-derivative}
\end{equation}
whereas an elementary joint transformation in
\eqref{eq:appendix-joint-exp} satisfies
\begin{equation}
    \left[
        \frac{\partial \boldsymbol{\mathcal{T}}_{J,j}}
             {\partial q_j}
        \boldsymbol{\mathcal{T}}_{J,j}^{-1}
    \right]^\vee
    =
    s_j\boldsymbol{\xi}_j.
    \label{eq:appendix-joint-derivative}
\end{equation}
For a coordinate not contained in a given local factor, the corresponding derivative is zero. Accordingly, after the fixed and joint-dependent factors have been arranged in their traversal order, the complete path transformation is obtained as
\begin{equation}
    \boldsymbol{T}_{\ell}^{\langle\sigma\rangle}
    =
    \prod_{k=1}^{n_{\ell}^{\langle\sigma\rangle}}
    \boldsymbol{\mathcal{T}}_{\ell,k}^{\langle\sigma\rangle}.
    \label{eq:appendix-path-product}
\end{equation}
If a generalized coordinate occurs in more than one local factor, each occurrence contributes separately to the complete-path derivative. When it occurs in only one local transformation, the summation in \eqref{eq:complete-path-derivative} consequently reduces to the corresponding single adjoint-transported term.

\section*{Acknowledgements}
We acknowledge the financial support of the Finnish Ministry of Education and Culture through the Intelligent Work Machines Doctoral Education Pilot Program (IWM VN/3137/2024-OKM-4).

\bibliographystyle{elsarticle-num}
\bibliography{ref}

\end{document}